%% file: arxiv.tex
\documentclass[twocolumn, conference]{IEEEtran}
\IEEEoverridecommandlockouts
\usepackage{cite}
\usepackage{amsmath,amssymb,amsfonts}
\usepackage{algorithmic}
\usepackage{graphicx}
\usepackage{multirow}
\usepackage{balance}
\usepackage{algorithm,algorithmic}
\usepackage{hyperref}
\usepackage{subfigure}
\hypersetup{hidelinks=true}
\usepackage{textcomp}

\begin{document}

\title{MTS-LOF: Medical Time-Series Representation Learning via Occlusion-Invariant Features}

\author{
    \IEEEauthorblockN{Huayu Li, Ana S. Carreon-Rascon, Xiwen Chen, Geng Yuan, and Ao Li}
    \thanks{This work was supported by grants from the National Heart, Lung, and Blood Institute (\#R21HL159661), and the National Science Foundation (\#2052528).}
    \thanks{Huayu Li, Ana S. Carreon-Rascon are with the Department of Electrical \& Computer Engineering at the University of Arizona, Tucson, AZ 85719 USA. (e-mail: hl459@arizona.edu, anascarreonr@arizona.edu)}
    \thanks{Xiwen Chen is with the School of Computing at Clemson University, Clemson, SC 29634 USA. (e-mail: xiwenc@g.clemson.edu)}
    \thanks{Geng Yuan is with the School of Computing (CS Department) at the University of Georgia,  Athens, GA 30602 USA. (e-mail: geng.yuan@uga.edu)}
    \thanks{Ao Li is with the Department of Electrical \& Computer Engineering and BIO5 Institute at The University of Arizona, Tucson, AZ 85719 USA. (e-mail: aoli1@arizona.edu)}
}
\maketitle

\begin{abstract}
Medical time series data are indispensable in healthcare, providing critical insights for disease diagnosis, treatment planning, and patient management. The exponential growth in data complexity, driven by advanced sensor technologies, has presented challenges related to data labeling. Self-supervised learning (SSL) has emerged as a transformative approach to address these challenges, eliminating the need for extensive human annotation. In this study, we introduce a novel framework for Medical Time Series Representation Learning, known as MTS-LOF. MTS-LOF leverages the strengths of contrastive learning and Masked Autoencoder (MAE) methods, offering a unique approach to representation learning for medical time series data. By combining these techniques, MTS-LOF enhances the potential of healthcare applications by providing more sophisticated, context-rich representations. Additionally, MTS-LOF employs a multi-masking strategy to facilitate occlusion-invariant feature learning. This approach allows the model to create multiple views of the data by masking portions of it. By minimizing the discrepancy between the representations of these masked patches and the fully visible patches, MTS-LOF learns to capture rich contextual information within medical time series datasets. The results of experiments conducted on diverse medical time series datasets demonstrate the superiority of MTS-LOF over other methods. These findings hold promise for significantly enhancing healthcare applications by improving representation learning. Furthermore, our work delves into the integration of joint-embedding SSL and MAE techniques, shedding light on the intricate interplay between temporal and structural dependencies in healthcare data. This understanding is crucial, as it allows us to grasp the complexities of healthcare data analysis. 
\end{abstract} 


\input{Content/introduction}

\input{Content/relatedworks}

\begin{figure}[!]
    \centering
    \includegraphics[width=.49\textwidth]{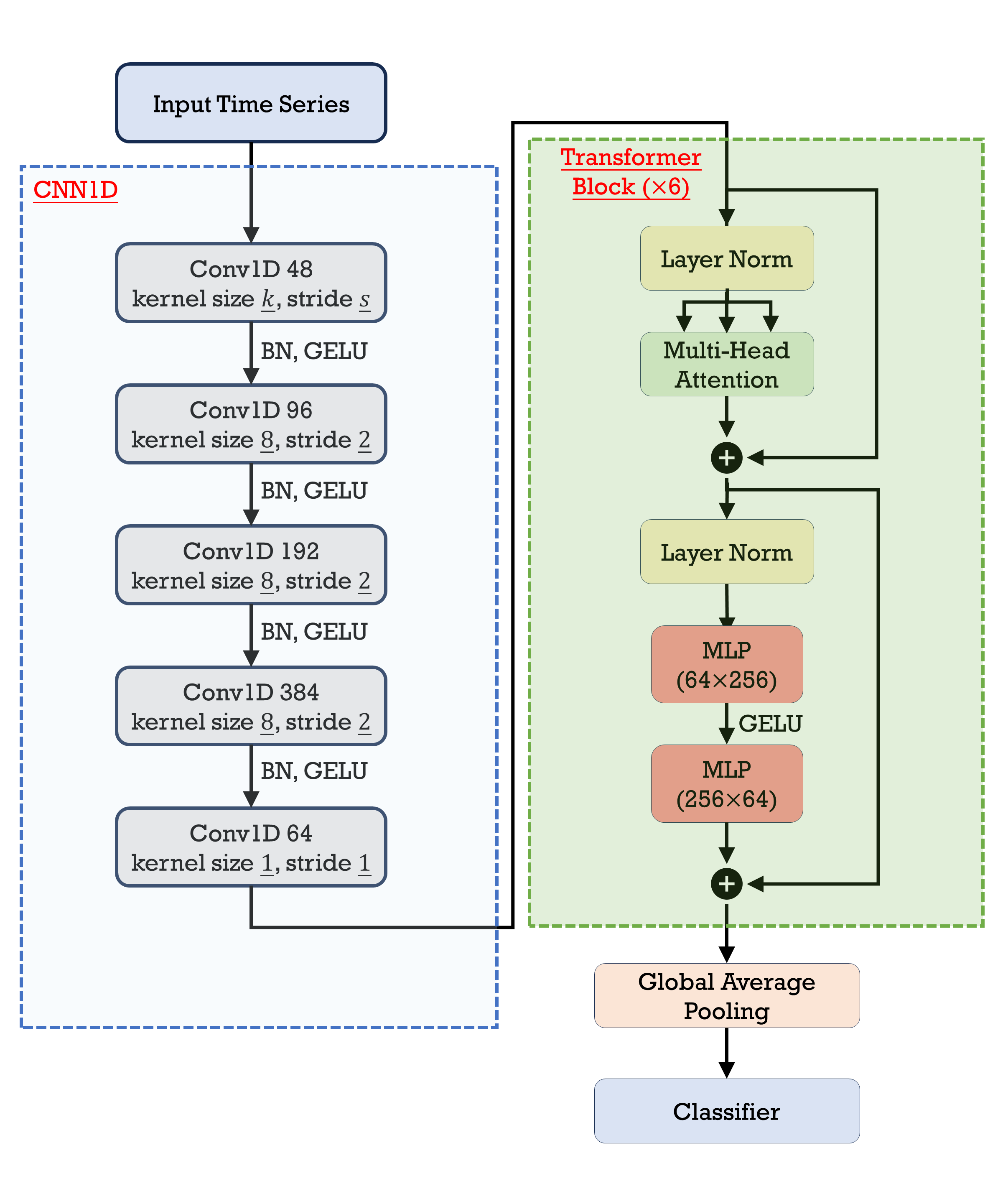}
    \caption{Illustration of the backbone network architecture. This figure provides an overview of the backbone network employed in our study, designed to effectively process multidimensional multivariate time series samples. The input time series undergoes a patching process using a CNN1D, followed by transformation through the transformer encoder to generate meaningful representations. The final representations are obtained from the outputs of the transformer encoder post a global average pooling layer. These representations are then input into a linear classifier to make the final predictions.}
    \label{fig:network}
\end{figure}

\input{Content/methods}
\begin{figure}[!]
    \centering
    \includegraphics[width=.45\textwidth]{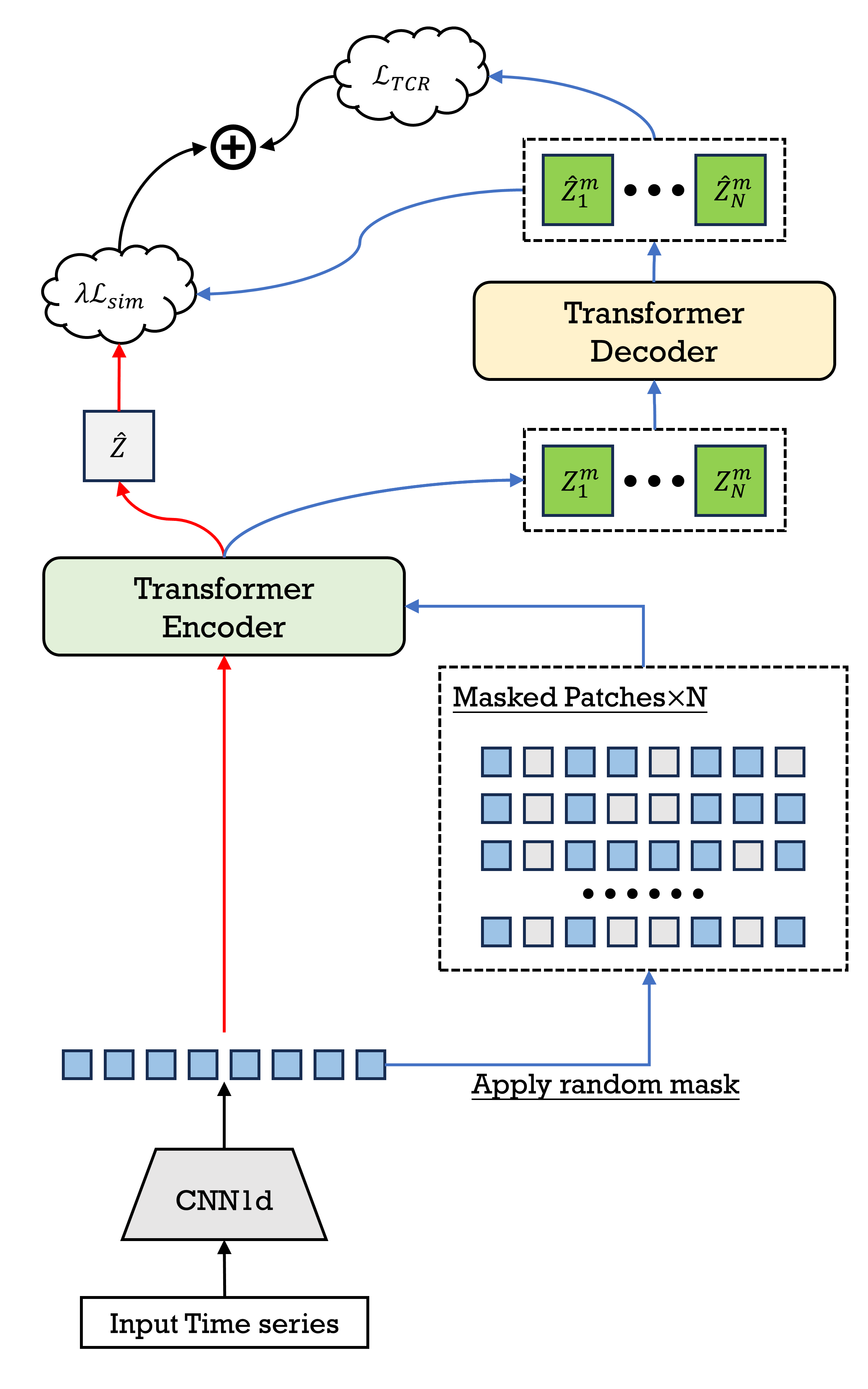}
        \caption{Illustration of the MTS-LOF framework workflow. The framework leverages Occlusion-Invariant Feature Learning (MAE) and Joint-Embedding SSL principles to enhance representation robustness. It employs multiple mask operations to enhance consistency in occlusion-invariant features, ensuring the model's effectiveness in the presence of occluded data. The similarity objective ($\mathcal{L}_{sim}$) measures the agreement between masked and unmasked representations, while covariance regularization ($\mathcal{L}_{TCR}$) is employed to mitigate representation collapse. A transformer decoder and positional embeddings contribute to comprehensive feature extraction. The hyperparameter $\lambda$ balances $\mathcal{L}_{sim}$ and $\mathcal{L}_{TCR}$.}
    \label{fig:algorithm}
\end{figure}

\input{tables/dataset_description}
\input{tables/main_results}

\begin{figure}[!]
    \centering
    \includegraphics[width=.45\textwidth]{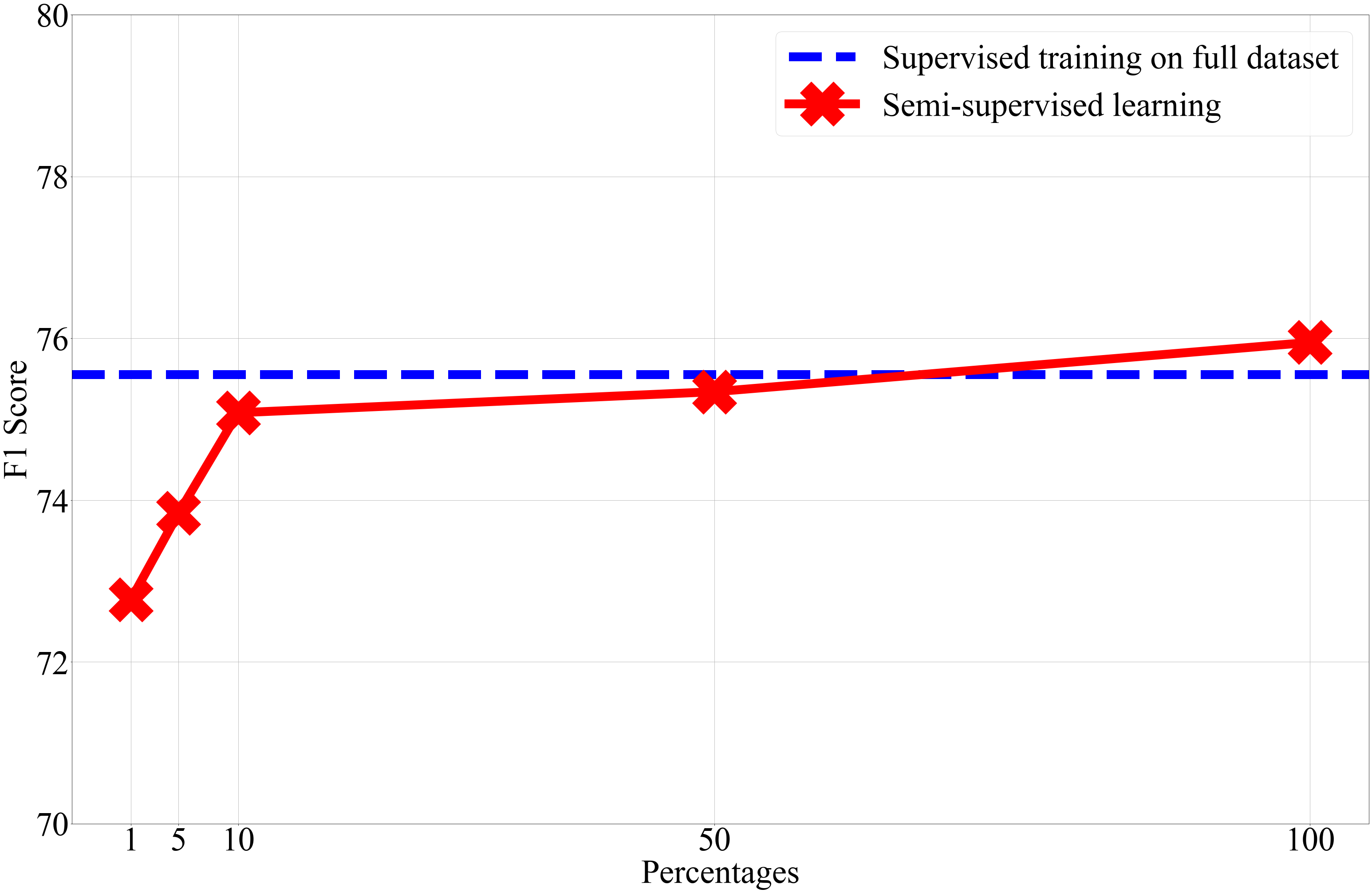}
    \caption{Fine-tuning the pretrained backbone with different fractions of labeled data from the Sleep-EDF dataset. The plot illustrates the performance of the MTS-LOF framework under semi-supervised learning conditions, comparing F1 scores obtained with 1\%, 5\%, 10\%, 50\%, and 100\% of randomly selected subsets of labeled data to the fully supervised learning result with 100\% labeled data. These results highlight the framework's adaptability and ability to leverage minimal labeled data effectively.}
    \label{fig:semi_supervised}
\end{figure}

\input{tables/shhs}
\input{tables/fd}

\begin{figure*}[!]
    \centering
    \subfigure[F1 score vs. number of masks.]{
        \includegraphics[width=.45\textwidth]{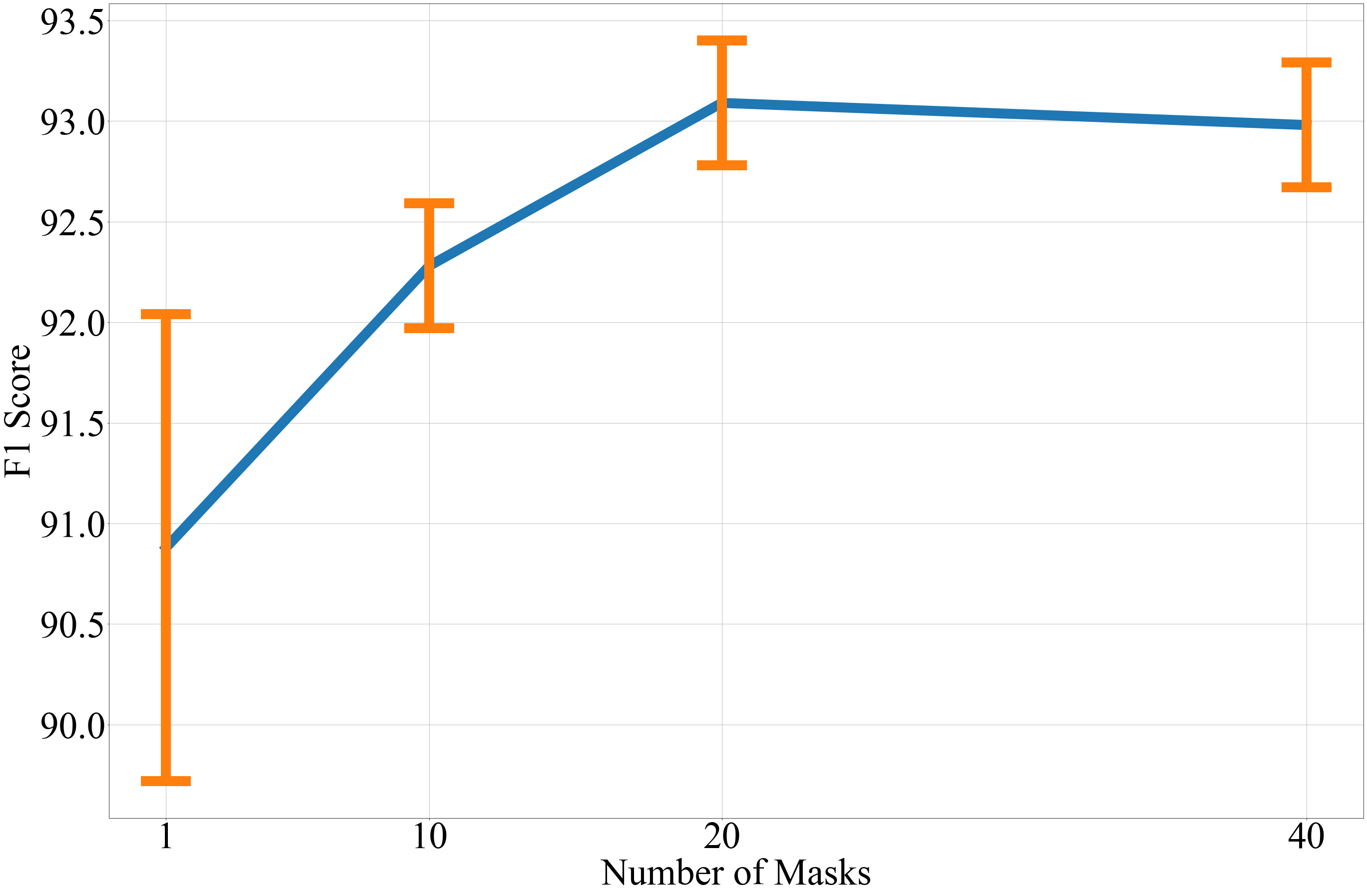}
    }
    \subfigure[F1 score vs. mask ratio.]{
        \includegraphics[width=.45\textwidth]{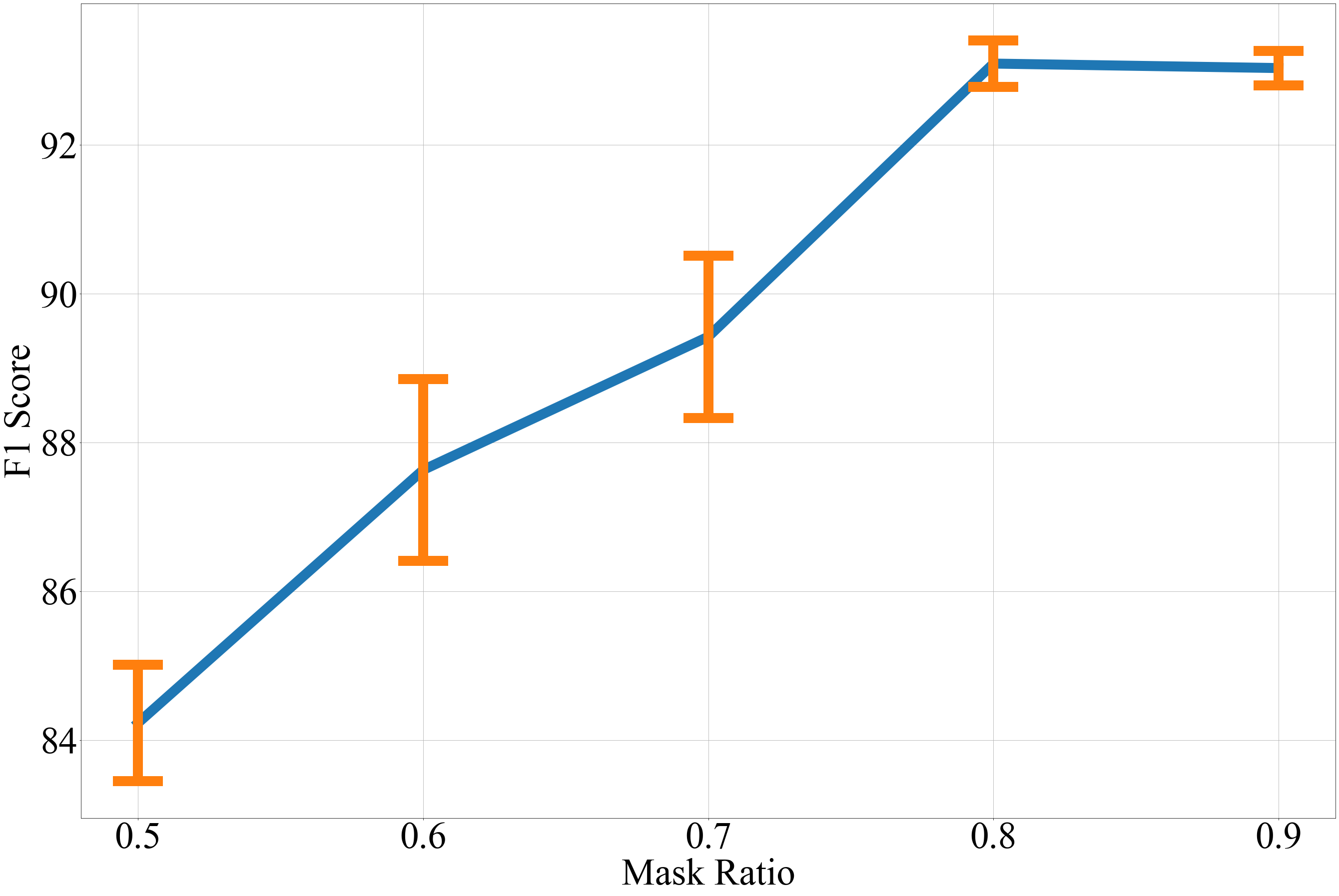}
    }
    \caption{Comparison of F1 scores under different combinations of hyperparameters in the ablation study using the HAR dataset. (a) Illustrates the relationship between F1 score and the number of masks while keeping the mask ratio constant at 0.8. (b) Shows the impact of varying mask ratios on the F1 score while maintaining a constant number of 20 masks. These findings provide insights into the sensitivity of the F1 score to these critical hyperparameters.}
    \label{fig:ablation}
\end{figure*}

\begin{figure*}
    \centering
    \subfigure[SSL on Epilepsy.]{
        \includegraphics[width=.3\textwidth]{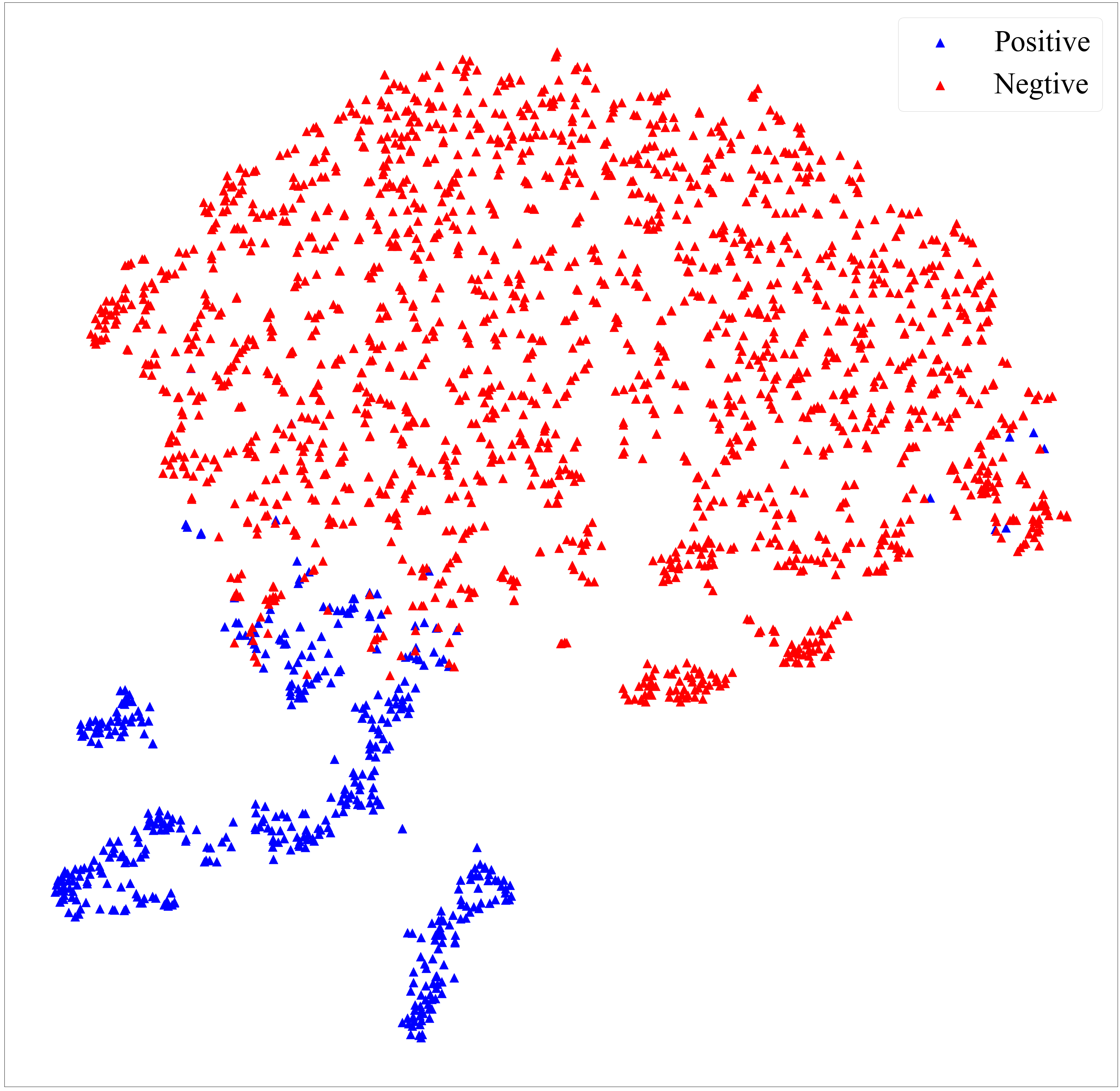}
    }
    \subfigure[Supervised training on Epilepsy.]{
        \includegraphics[width=.3\textwidth]{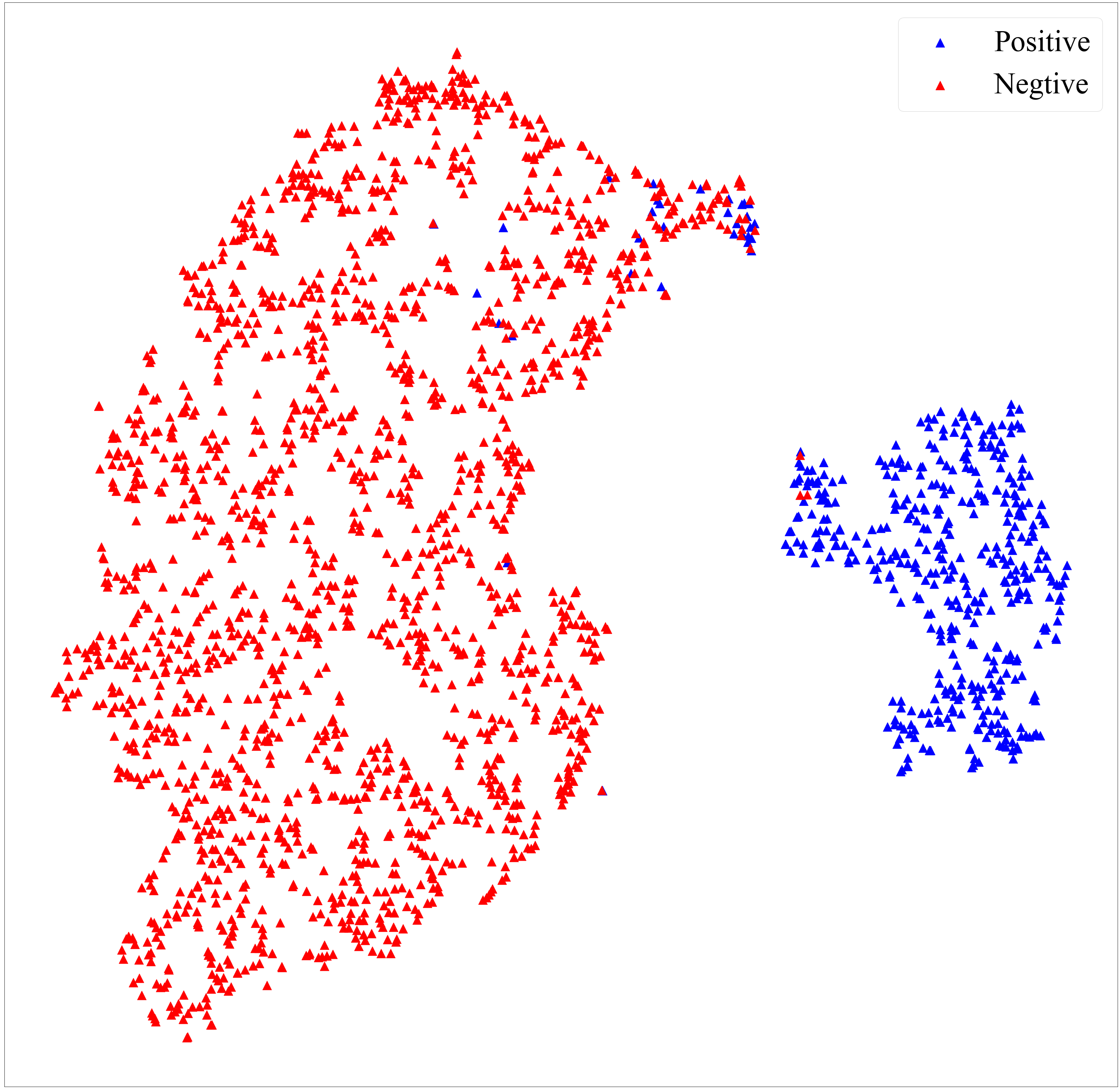}
    }
    \subfigure[Finetuning on Epilepsy.]{
        \includegraphics[width=.3\textwidth]{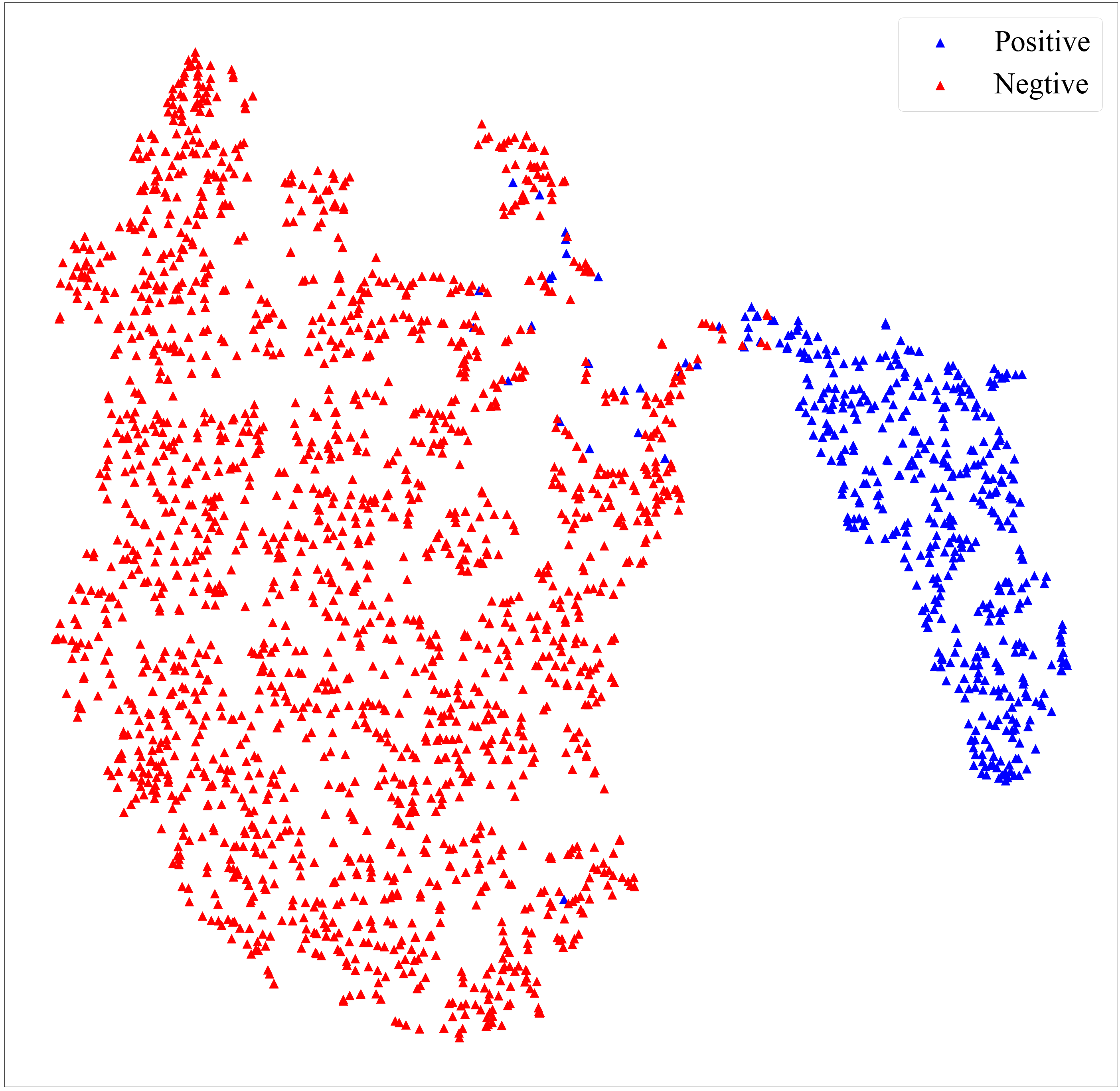}
    }
    \subfigure[SSL on HAR.]{
        \includegraphics[width=.3\textwidth]{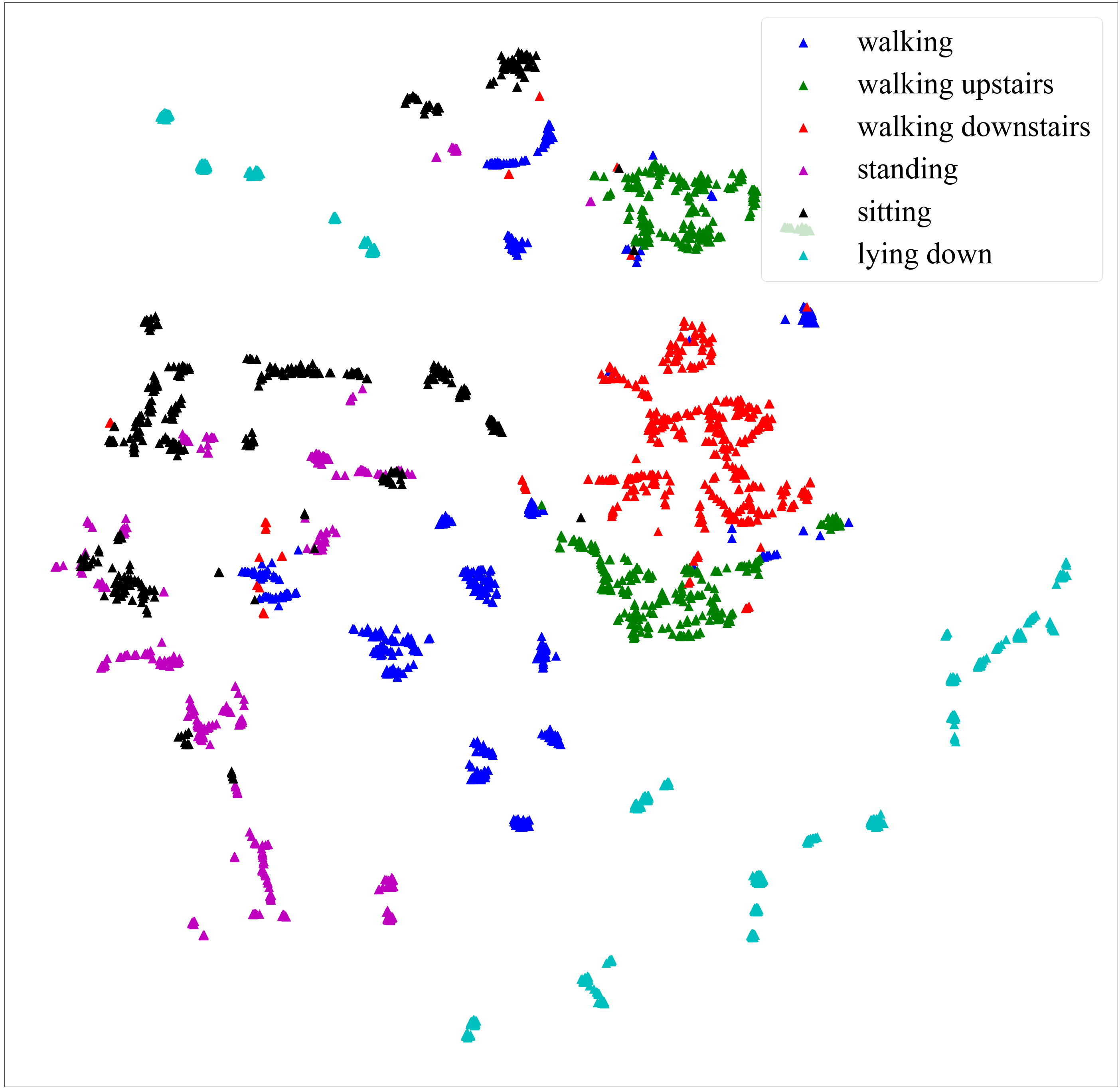}
    }
    \subfigure[Supervised training on HAR.]{
        \includegraphics[width=.3\textwidth]{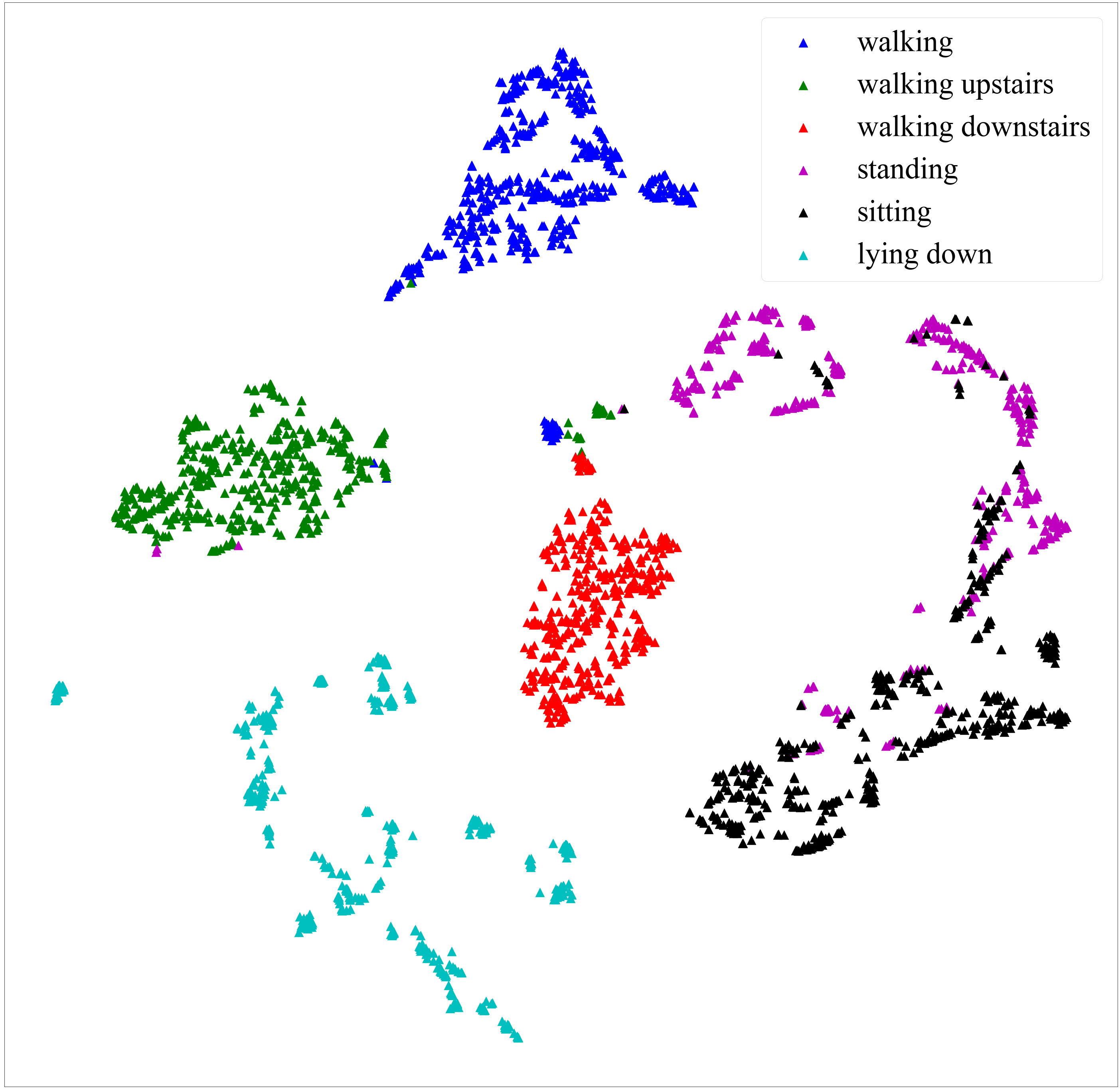}
    }
    \subfigure[Finetuning on HAR.]{
        \includegraphics[width=.3\textwidth]{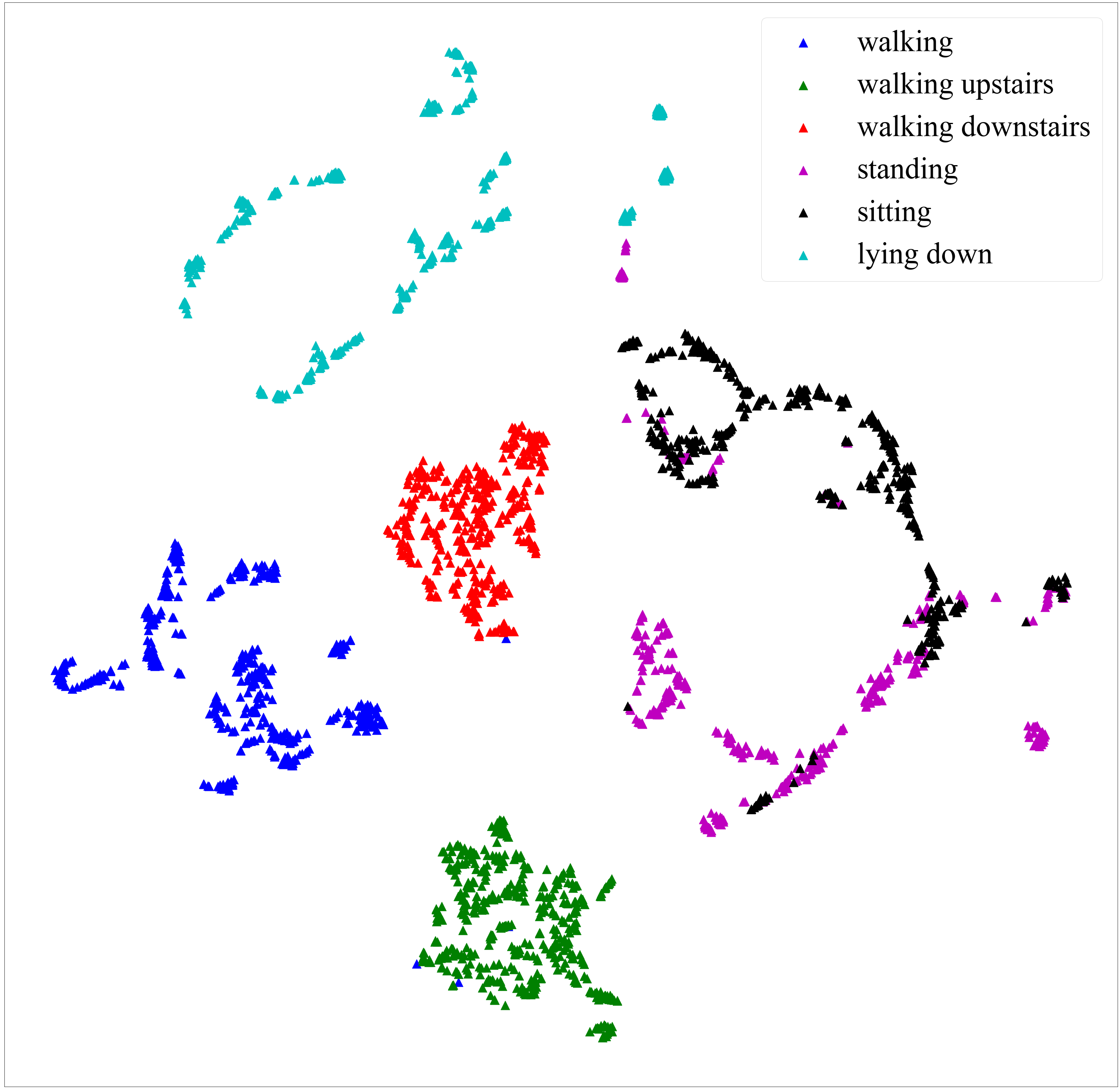}
    }
    \subfigure[SSL on Sleep-EDF.]{
        \includegraphics[width=.3\textwidth]{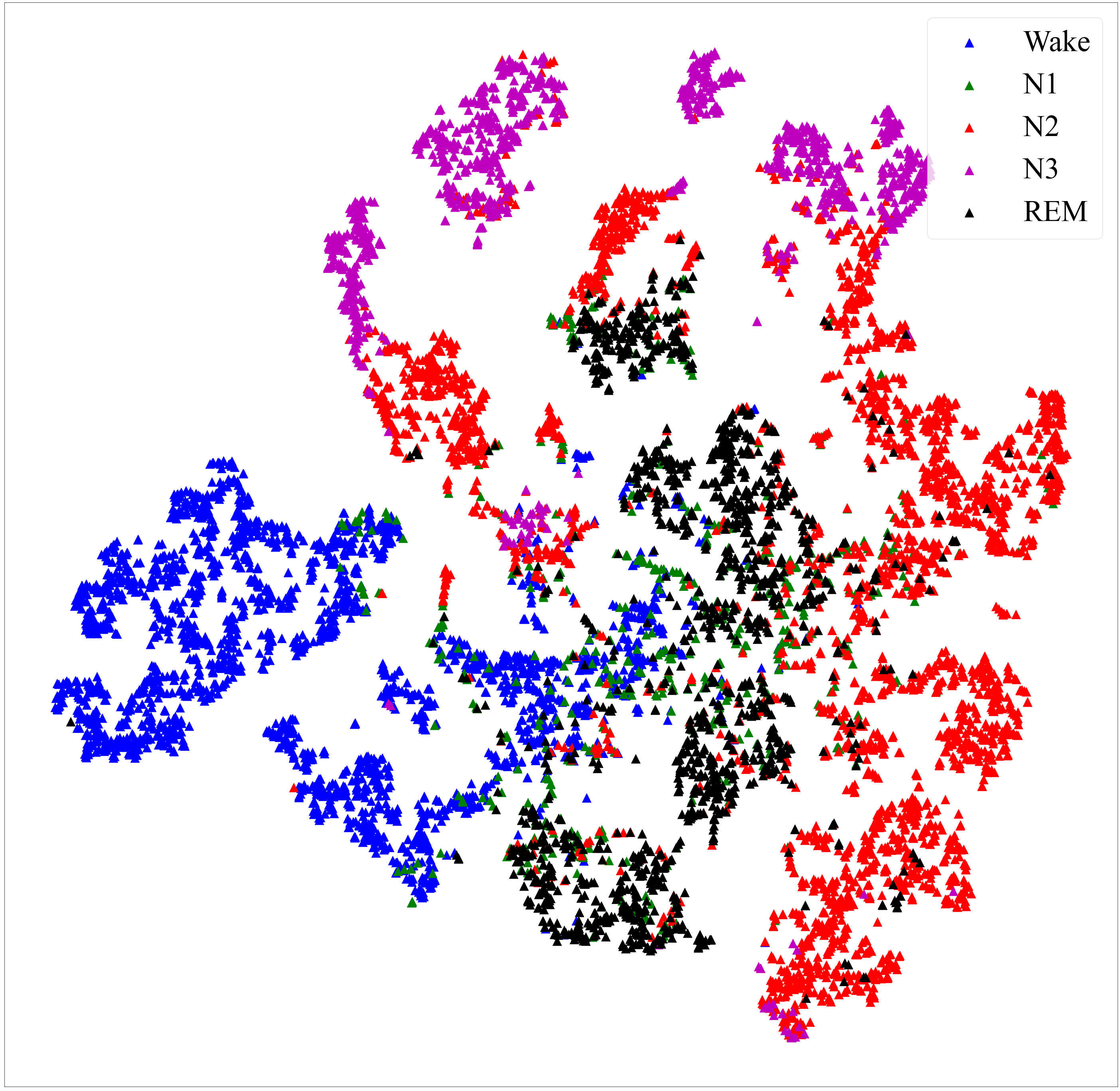}
    }
    \subfigure[Supervised training on Sleep-EDF.]{
        \includegraphics[width=.3\textwidth]{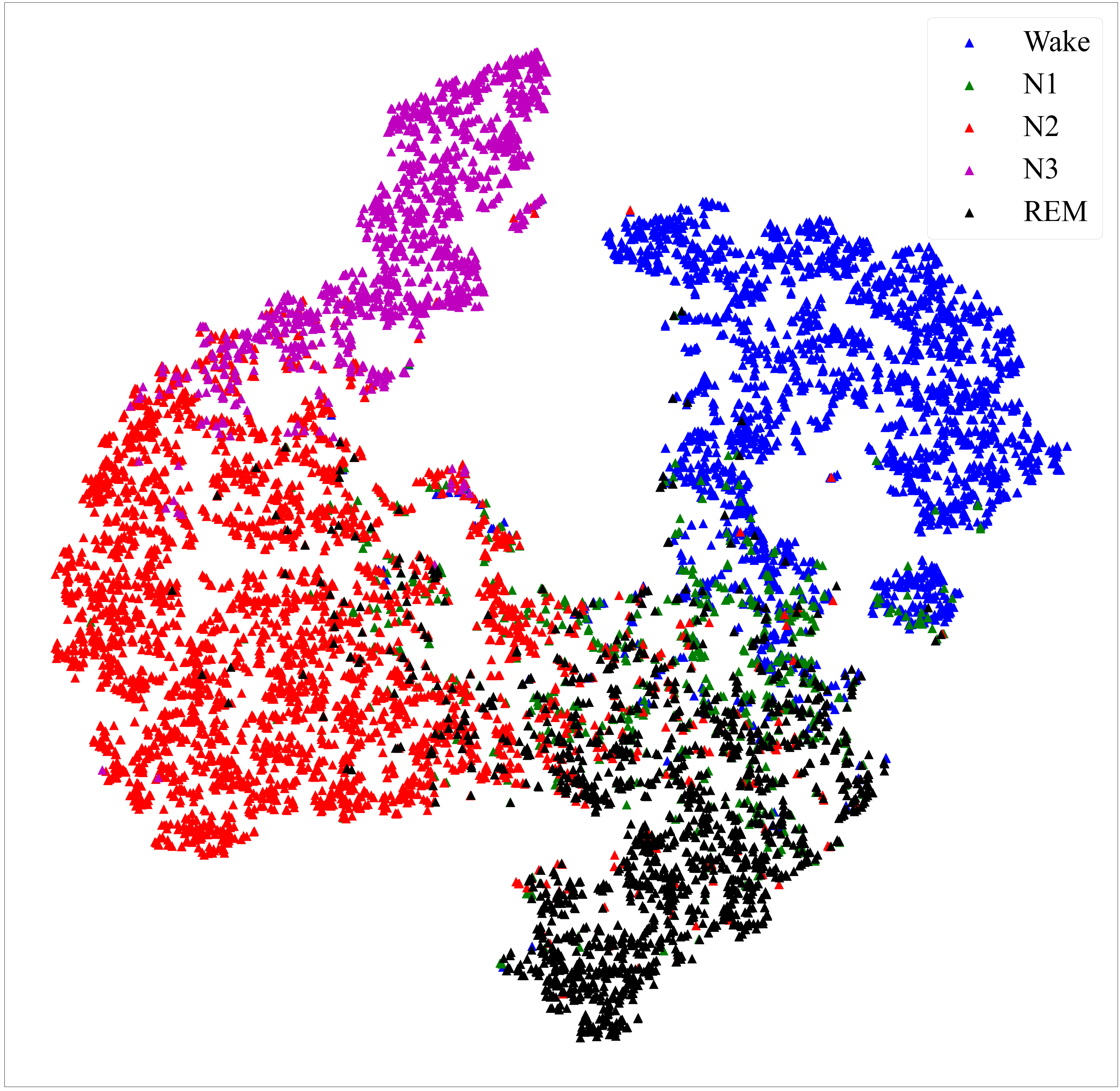}
    }
    \subfigure[Finetuning on Sleep-EDF.]{
        \includegraphics[width=.3\textwidth]{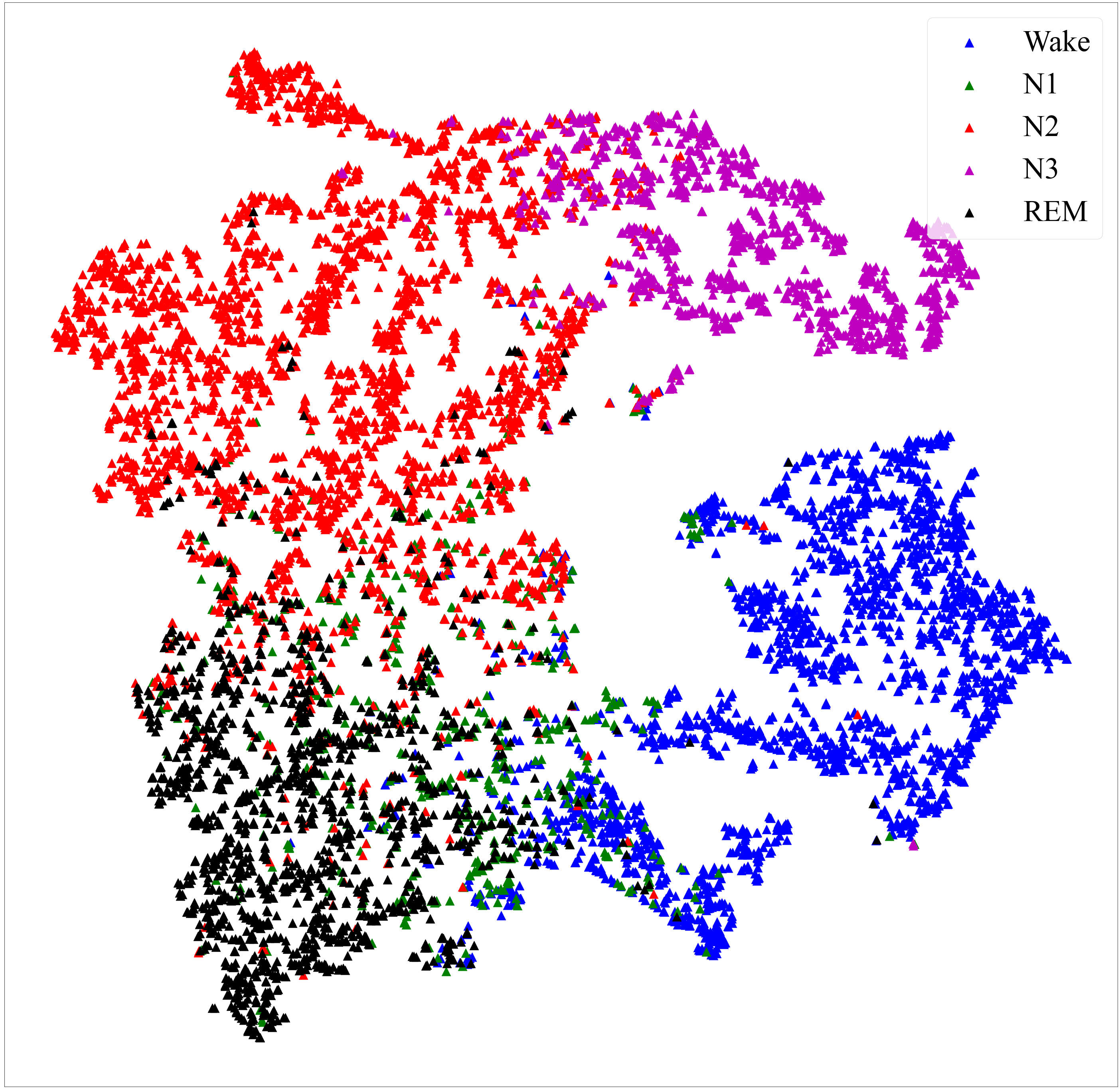}
    }
    \caption{t-SNE visualizations of learned representations for Epilepsy, HAR, and Sleep-EDF datasets. The subfigures display the effects of different training methodologies: SSL, supervised training, and 5\% fine-tuning on each dataset.}
    \label{fig:tsne}
\end{figure*}

\input{Content/expirements}

\input{Content/discussion}

\balance
\bibliography{ref}
\bibliographystyle{IEEEtran}




\end{document}

%% file: Content/introduction.tex
\section{Introduction}

Medical time series data, comprising physiological signals, human activities, and other time-stamped data, is crucial across various clinical settings and home care scenarios, guiding clinical decision-making and patient management. The analysis and interpretation of these complex temporal data streams are pivotal for enhancing our comprehension of health and facilitating timely interventions. Recently, the exponential growth in both the volume and complexity of medical time series data has been propelled by advanced sensor technologies and enhancements in electronic health records. Consequently, the automatic classification of medical time series data represents a fundamental challenge in the development of intelligent clinical decision support systems. Typically, classification models are trained using supervised learning on labeled data pairs annotated by human experts. Nevertheless, the continuous generation of vast volumes of time series data from various sensors renders human annotation impractical and inconsistent with the cost-effectiveness principle \cite{ching2018opportunities}.

In recent years, self-supervised learning (SSL) techniques have emerged as a transformative approach across various domains, including computer vision\cite{noroozi2016unsupervised, gidaris2018unsupervised, oord2018representation,chen2020simple,he2020momentum,hjelm2018learning}, natural language processing \cite{devlin2018bert,brown2020language,radford2018improving,liu2019roberta}, and time series \cite{franceschi2019unsupervised,eldele2021time,sarkar2020self,yue2022ts2vec,mohsenvand2020contrastive}. Supervised learning methods, which rely on extensive labeled data and domain expertise for training, face challenges related to the availability of large annotated datasets.	Yet, obtaining labeled data is impractical in numerous scenarios due to factors like time constraints, cost considerations, or privacy concerns. In response to these challenges, SSL techniques have gained prominence as a game-changing innovation.	They provide a versatile alternative that empowers models to autonomously derive meaningful representations directly from raw data, eliminating the need for extensive reliance on labeled samples. As a result, SSL has opened new avenues for knowledge discovery, enabling more efficient, cost-effective, and privacy-conscious solutions to complex data analysis tasks.	

The invariance-based methods (also called joint-embedding SSL) contain two branches, Contrastive learning \cite{hadsell2006dimensionality,oord2018representation, bachman2019learning, caron2020unsupervised, chen2020simple, he2020momentum} and Non-Contrastive learning \cite{grill2020bootstrap, bardes2021vicreg, zbontar2021barlow, chen2021exploring}, that share the same core idea, which is to learn a representation invariant to augmentations of the same image instance. It means that the learned representation of two different augmentations of the same image should be close, and the representation space shall not be a collapsed trivial one (e.g. all representations collapse to a constant). To overcome those issues, contrastive learning methods create a dichotomy between positive and negative samples in a latent space, thereby encouraging the model to pull similar samples together while pushing dissimilar ones apart. In contrast, Non-Contrastive learning methods learn to produce similar embeddings for different views of the same image, and penalize the embeddings with regularization or stop gradients to enforce a non-collapsing solution. Recent time series representation learning algorithms \cite{mohsenvand2020contrastive,eldele2021time,yue2022ts2vec} are developed based on contrastive learning and achieved outstanding improvements than different baselines \cite{saeed2019multi, sarkar2020self}. Compared to SSL in computer vision tasks, time series data inherently possess temporal dependencies, a defining characteristic that is not adequately addressed by traditional image-based contrastive learning methods. Also, many augmentation techniques commonly employed in image-based contrastive learning, such as color distortion, do not seamlessly translate to time-series data, posing significant challenges to adapting these methods effectively.

Beneficial of the advancement of transformer architecture \cite{vaswani2017attention} in computer vision \cite{dosovitskiy2020image}, Masked Autoencoder (MAE) \cite{he2022masked}, and its broader concept, Masked Image Modeling (MIM) \cite{dosovitskiy2020image, bao2021beit}, are at the forefront of innovative approaches within the fields of computer vision, delivering robust methods for visual representation learning. These methodologies share a common foundational principle: the strategic concealment or "masking" of specific regions within an image, feature space, or patchified image. Through this selective occlusion, models are guided to scrutinize intricate patterns, complete missing information, and ultimately, provide valuable insights into the underlying structure of visual content. Much like the strides made in computer vision tasks, the work by Zerveas et al. \cite{zerveas2021transformer} presents an exciting development in the field of time series data analysis. They introduce a groundbreaking framework for multivariate time series representation learning based on the principles of MAE. This pioneering effort marks a significant leap forward as it represents the first direct application of MAE techniques to the domain of time series data analysis.

While time series contrastive learning and time series MAE methods have undeniably ushered in significant advancements in the domain of time series representation learning, a vital question emerges: Can we amalgamate the valuable insights gleaned from MAE with contrastive learning algorithms to craft invariant representations that encompass an even more profound wealth of contextual information? This question lies at the heart of a relentless pursuit to amplify the capabilities of time series representation learning. Contrastive learning and MAE techniques have independently demonstrated their prowess by tackling various facets of feature extraction and data representation. Contrastive learning excels in the discernment of discriminative patterns within time series, while MAE exhibits a remarkable aptitude for capturing the latent dependencies and nuanced context embedded in temporal data. 

The significance of this potential integration cannot be overstated. It holds the promise of furnishing time series representation learning with a new level of sophistication and depth, enabling it to yield more comprehensive and context-rich representations. More importantly, such an integration benefits more in the medical time series domain where the temporal and the structural dependency share almost the same importance. In clinical decision-making procedures, physicians base their diagnoses on the local patterns of the time series data, such as interictal spikes in EEGs. Conversely, some diagnoses depend more on the long term temporal behaviors of the data, such as sleep architecture. 

In this paper, we envision the creation of a novel framework that could yield representations by integrating the strengths of these two approaches. We propose a \textbf{M}edical \textbf{T}ime-\textbf{S}eries representation \textbf{L}earning framework via \textbf{O}cclusion-invariant \textbf{F}eature (MTS-LOF).  The proposed MTS-LOF framework employs a simple yet efficient multi-masking strategy without specific data augmentation to create multiple views of the patches of data. Then, the MTS-LOF framework learns the occlusion-invariant feature by minimizing the discrepancy between each masked patch and the complete data. Experiments are conducted on different medical time series data and the results demonstrate the performance of MTS-LOF outperforms different baselines. In summary, the main contributions of this work are as follows:
\begin{itemize}
    \item \textbf{MTS-LOF Framework}: We present the MTS-LOF framework, a novel paradigm for medical time series representation learning. Unlike other frameworks that rely on specific data augmentation, this framework offers a simple yet highly effective approach to capture rich contextual information within medical time series data.
    \item \textbf{Integration of Joint-Embedding SSL and MAE}: MTS-LOF is one of the first frameworks that integrates the advantages of both contrastive learning and MAE techniques in medical time series representation learning, capturing the intricate interplay between temporal and structural dependencies within the data. This fusion underscores the potential for enhanced representation learning in healthcare applications.
    \item \textbf{Performance Evaluation on Diverse Medical Datasets}: We conducted extensive experiments on several medical datasets, including Epilepsy Seizure Prediction\cite{andrzejak2001indications}, Human Activity Recognition\cite{anguita2013public}, Sleep-EDF\cite{goldberger2000physiobank}, and Sleep Heart Health Study\cite{quan1997sleep}. The experimental results demonstrate that our framework significantly outperforms state-of-the-art methods. This framework has great potential for healthcare applications in clinical settings and home care.  
\end{itemize}


%% file: Content/relatedworks.tex
\section{Related works}

\subsection{Joint-Embedding Self-Supervised Learning} 
Joint-Embedding SSL is a prominent paradigm in representation learning, dedicated to discerning similarities and dissimilarities among data point pairs.	It primarily involves maximizing the similarity between augmented views from identical inputs (referred to as "positive" samples) and minimizing the similarity between augmented views from distinct inputs (referred to as "negative" samples). A primary challenge in Joint-Embedding SSL is the potential risk of representation collapse. In such cases, the model consistently generates identical outputs, making the learned representations uninformative. In the realm of contrastive learning, the methodologies frequently incorporate positive samples and negative samples. The overarching objective is to bring positive samples closer to the anchor sample within an embedding space while simultaneously pushing negative samples farther apart. Contrastive learning techniques, including Contrastive Predictive Coding (CPC) \cite{oord2018representation}, Simple Contrastive Learning Framework (SimCLR) \cite{chen2020simple}, and Momentum Contrast (MoCo) \cite{he2020momentum}, have demonstrated substantial progress. Unlike contrastive learning methods, non-contrastive approaches in Joint-Embedding SSL depart from the conventional utilization of negative samples. These techniques employ diverse strategies to mitigate the risk of representation collapse. For example, Bootstrap Your Own Latent (BYOL) \cite{grill2020bootstrap} and Simple Siamese representation learning (SimSiam) \cite{chen2021exploring} employ gradient stopping techniques and predictor modules to avert representation collapse. Furthermore, Variance-Invariance-Covariance Regularization (VICReg) \cite{bardes2021vicreg} introduces covariance regularization as a means to penalize learned representations, thus mitigating the risk of collapse and enhancing the stability of embeddings.	

\subsection{Masked Auto Encoder and Masked Image Modeling}
Within the domain of SSL, masked language modeling (MLM) has emerged as a prominent method in Natural Language Processing (NLP) \cite{devlin2018bert, liu2019roberta, brown2020language}. MLM and its auto-regressive variations \cite{brown2020language} have transformed NLP, enabling the training of extensive language models that excel in a wide range of language comprehension and generation tasks. These methodologies entail predicting masked tokens within sentences or sentence pairs/triplets, harnessing extensive training data to attain remarkable performance. Concurrently, Masked Image Modeling (MIM) has advanced in parallel with MLM. These techniques encompass the deliberate masking of particular regions within images or feature spaces, directing models to discern intricate patterns, imputing missing information, and ultimately extracting valuable insights regarding the underlying structure of visual content. A seminal contribution in this domain is the context encoder approach \cite{pathak2016context}, which masks regions within original images and predicts the missing pixels, laying the foundation for subsequent advancements. Recent studies \cite{chen2020generative, dosovitskiy2020image, bao2021beit} have delved into the utilization of MIM for pretraining vision Transformers \cite{dosovitskiy2020image, chen2021empirical}. Additionally, Masked Autoencoder (MAE) \cite{he2022masked} introduced a straightforward and scalable self-supervised learning approach for computer vision. This method involves masking random patches of the input image and subsequently reconstructing the missing pixels. Recent work proposed by Kong et al. \cite{kong2023understanding} reveals the insight behind MIM is encouraging the network to learn the learning occlusion invariant feature.

\subsection{Self-supervised Learning for Time-Series}
SSL techniques have significantly impacted the field of time-series data analysis. Researchers have earnestly explored the adaptation of SSL approaches to capture the intricate temporal dependencies, patterns, and representations inherent in sequential data. Before the emergence of time-series contrastive learning, the predominant method utilized multitask learning \cite{saeed2019multi, sarkar2020self} to obtain time-series representations. These methods typically involved applying various transformations to the original time series to construct pretext tasks and pseudo labels. Subsequently, models were trained to recognize the specific transformations that were applied. However, motivated by the remarkable successes observed in other domains, contrastive learning has made inroads into the domain of time-series representation learning. For instance, SimCLR \cite{chen2020simple} has been extended to support EEG representation learning \cite{mohsenvand2020contrastive}. Contrastive Predictive Coding (CPC) \cite{oord2018representation} has employed predictive modeling in latent spaces to acquire time-series representations, showcasing competence in speech recognition tasks. The TS-TCC framework \cite{eldele2021time} has introduced innovative temporal and contextual contrasting modules founded on transformer blocks \cite{vaswani2017attention}, enhancing the acquisition of discriminative representations. Additionally, TS2Vec \cite{yue2022ts2vec} has presented a hierarchical contrasting framework with the capability to capture multi-scale temporal and instance dependencies within time series data. In addition to temporal contrastive learning, the direct application of MAE in the context of time series is put forth in \cite{zerveas2021transformer}.

%% file: Content/methods.tex
\section{Proposed Methods}

\subsection{Backbone Model Structure}
We begin by detailing the backbone network utilized in our study. Specifically, when dealing with a collection of multidimensional multivariate time series samples, each training sample denoted as $X\in\mathbb{R}^{t\times m}$ constitutes a sequence of $m$ univariate series represented as $x\in \mathbb{R}^{t}$, where $t$ signifies the time step. As illustrated in Figure. \ref{fig:network}, our approach diverges from prior work \cite{zerveas2021transformer}, which employed a linear projection of each time step from the original time series into a $d$-dimensional vector space, treating them as tokens, using $\hat{X}=WX+b$, with $W\in\mathbb{R}^{m\times d}$ and $b\in\mathbb{R}^d$. Instead, we adopt a strategy where we initially generate patches as tokens by processing the time series through several convolutional layers. To streamline the discussion, we implement a modified version of the CNN1D architecture from TS-TCC \cite{eldele2021time}. The configuration of the initial convolutional layer varies depending on the dataset, with distinct kernel sizes ($k$) and strides ($s$). Subsequently, we stack three convolutional layers with fixed kernel size $8$ and stride $2$. These layers efficiently downsample the original time series by a factor of $8\times s$. Finally, a single convolutional layer with kernel size $1$ and stride $1$ reduces the feature channel dimensions to match the $d$-dimensional input required by the transformer encoder. Each convolutional layer is followed by a batch normalization (BN) \cite{ioffe2015batch} and Gaussian Error Linear Units (GELU) \cite{hendrycks2016gaussian} beside the last one. Several factors motivate this choice. Firstly, akin to the concept of patching in vision transformers (ViT) \cite{dosovitskiy2020image}, we consider a single time step as equivalent to a single pixel, lacking semantic meaning akin to words in a sentence. Thus, it becomes imperative to extract local semantic information for analyzing their connections. Secondly, as exemplified in \cite{xiao2021early}, early convolutional layers can enhance the stability and convergence of the transformer during training. Lastly, this design facilitates further modifications and extensions of the backbone network, enabling the replacement of the simple CNN1D with more complex architectures \cite{eldele2021attention}.

After generating the patches through the convolutional layers, the tokens $\hat{X}\in\mathbb{R}^{p\times d}$, where $p = \frac{t}{8s}$ represents the number of patches, are encoded by adding a fixed 1D cosine positional embedding denoted as $\text{P}$, defined as follows:
\begin{align}
    \text{P}_{i,2\cdot j} = \sin(\frac{i}{10000^{2j/d}}),\quad \text{P}_{i,2\cdot j +1} = \cos(\frac{i}{10000^{2j/d}}).
\end{align}
Here, $\text{P}_{i,j}$ signifies the value of the embedding for the $i$-th position and the $j$-th dimension. Typically, $d$ matches the dimension of the Transformer encoder's input embeddings. This positional encoding method enhances the model's capability to capture sequential order and relationships within the time series data, which are crucial for various time-series analysis tasks. It ensures that the model can differentiate between different positions in the sequence, and when combined with the patch-based tokenization, it allows the transformer encoder to effectively process the data for downstream tasks. Subsequently, we feed these encoded tokens into the Transformer encoder. The Transformer encoder employs multi-head attention, where each attention head, denoted as $h=1,\dots,H$, independently processes the input tokens $\hat{X}$. Each head transforms the input into query matrices $Q_h$, key matrices $K_h$, and value matrices $V_h$, achieved through weight matrices $W_Q$, $W_K$, and $W_V$, respectively. Once the query, key, and value matrices are obtained, a scaled dot-product attention mechanism computes the attention output $O_h$. The softmax function is utilized to calculate attention weights, which determine the level of attention each patch pays to others. The calculation for $O_h$ is defined as:
\begin{align}
    O^{\dagger}_h = \text{Attention}(Q_h,K_h,V_h) = \text{softmax}(\frac{Q_{h}K_{h}^{\dagger}}{\sqrt{d_k}})V_{h},
\end{align}
where the $\sqrt{d_k}$ term refers to the dimension of the key vectors and is employed to scale the dot product to prevent it from becoming excessively large. The multi-head attention block incorporates Layer Normalization layers \cite{ba2016layer} to normalize the data and stabilize training. Additionally, a feed-forward network with residual connections is employed to capture intricate patterns within the data. Consequently, the outcome of the Transformer encoder's operations is a representation denoted as $Z\in\mathbb{R}^{p\times d}$. Finally, a global average pooling layer is applied to obtain the final representation $\hat{Z}\in\mathbb{R}^{d}$, followed by a linear head for predicting results $Y\in\mathbb{R}^{c}$, where $c$ is the number of classes.

\subsection{Occlusion-Invariant Feature Learning}
We now delve into the concept of Occlusion-Invariant Feature Learning and motivation behind the proposed MTS-LOF framework. To establish a comprehensive understanding, it is essential to revisit the MAE objective, a pivotal component of our methodology. The primary purpose of this objective is to predict the original pixel values of the masked patches/tokens $\hat{X}\odot M$. Let it be known that we represent $M \in \mathbb{R}^p$, where each $m_i$ within $M$ is assigned the value of $1$ for masked tokens and $0$ for unmasked tokens, effectively delineating the regions of the time series data that are concealed and revealed. This MAE objective, working in tandem with a transformer decoder denoted as $d_\phi(\cdot)$ and the latent representation $Z^m$ obtained from the unmasked patches $\hat{X}\odot(1-M)$, is formally articulated as follows:
\begin{align}
    \nonumber
    &\mathcal{L}_{MAE} = \left \| d_\phi(Z^m)\odot M- \hat{X}\odot M\right \|^2 \\
    &\approx \left \| d_\phi(Z^m)- \hat{X}\right \|^2, \quad Z^m = f_\theta(\hat{X}\odot(1-M)).
\end{align}
In this equation, $\mathcal{L}_{MAE}$ quantifies the mean absolute error between the predicted values obtained from the decoder and the actual values of the masked tokens. Fundamentally, the MAE objective encapsulates the idea of reconstructing the entirety of the input data, even when faced with incomplete or occluded information. It is worth noting that in an ideal scenario, with an over-parameterized neural network, the network can memorize all seen training samples. This implies that with the latent representation $Z$ of the complete input tokens $\hat{X}$, we have:
\begin{align}
    d_\phi(Z) = \hat{X},\quad Z = f_\theta(\hat{X}).
\end{align}
Consequently, we can reformulate the MAE objective with a distance function $\mathcal{D}$:
\begin{align}
    \mathcal{L}_{MAE} \approx \mathcal{D}(Z, Z^m) = \left \| Z^m- Z\right \|^2,
\end{align}
which signifies that the primary objective of MAE training is to acquire the Occlusion-Invariant Feature. 

Moreover, it is essential to recall the fundamental concept of Joint-Embedding SSL, which is centered on the objective of minimizing the distance between two latent representations derived from augmented views of the same input. This objective can be generally formulated as follows:
\begin{align}
    &\mathcal{L}_{Inv} = \mathcal{D}(Z_1, Z_2), \\
    &Z_1 = f_\theta(\mathcal{A}_{1}(\hat{X})), \quad Z_2 = f_\theta(\mathcal{A}_{2}(\hat{X})),
\end{align}
where we denote the two augmentations as $\mathcal{A}_{1}$ and $\mathcal{A}_{2}$. Importantly, it should be noted that the distance function can be both parametric and non-parametric. It means it can encompass simple functions such as mean square error and cosine similarity, or it can incorporate projection heads or predictor networks, as is common in Joint-Embedding SSL settings. We can establish a connection between Joint-Embedding SSL and the MAE framework by assuming that $\mathcal{A}{1}$ preserves the original input while $\mathcal{A}{2}$ involves random masking operations. Consequently, we can define the primary similarity objective of the proposed MTS-LOF Framework:
\begin{align}
    &\mathcal{L}_{sim} = \mathcal{D}(Z, Z^m),\\
    &Z^m = f_\theta(\hat{X}\odot(1-M)), \quad Z = f_\theta(\hat{X}).
\end{align}
This objective captures the essence of the MTS-LOF framework, aiming to minimize the distance between the original representation $Z$ and the representation $Z^m$ obtained after masking a portion of the input data. This formulation bears a resemblance to non-contrastive learning algorithms, which are prone to the risk of representation collapse. To mitigate this risk, a covariance regularization technique is employed to prevent representation collapse. Specifically, the latent representation $Z^m$ of the masked patches is penalized using Total Coding Rate (TCR), which is formulated as:
\begin{align}
    \mathcal{L}_{TCR}(Z) = \frac{1}{2}\log\det(I+\frac{d}{b\epsilon^2}ZZ^\dagger)
\end{align}
where $b$ represents the batch size, $d$ signifies the dimensionality of the representation, and $\epsilon>0$ is a chosen size of distortion.

In practical implementation, the algorithm is supposed to effectively capture features that are consistently occlusion-invariant. To achieve this, we apply multiple mask operations to the patches denoted as $\hat{X}$. This operation is defined as follows:
\begin{align}
\nonumber
    \hat{X}_1, \dots, \hat{X}_N &= \hat{X}\odot(1-M_1), \dots,\hat{X}\odot(1-M_N),\\
    &\text{where}\quad M_i \neq M_j \quad \text{for} \quad i\neq j.
\end{align}
where $N$ refers to the number of masks. Subsequently, the similarity objective is employed to compare the representations of the unmasked patches with each representation of the masked patches:
\newcommand{\lnorm}[1]{\frac{#1}{\left\lVert{#1}\right\rVert _2}}
\begin{align}
    \mathcal{L}_{sim}=-\frac{1}{N}\sum_{i=0}^{N} \lnorm{\hat{Z}}{\cdot}\lnorm{\hat{Z}_i^m},
\end{align}
where we use the negative cosine similarity to measure the distance between two representation vectors in this context. It's important to note that we calculate $\mathcal{L}_{sim}$ using the globally pooled representations $\hat{Z}$ and $\hat{Z}_i^m$, as well as the covariance regularization $\mathcal{L}_{TCR}$. A transformer decoder is employed to decode the unmasked patches encoded in the previous step, along with all the masked tokens. To provide positional awareness, position embeddings are added so that each token can determine its respective position. It's worth mentioning that each masked token is shared and can learn a vector. Visual demonstration of the entire workflow of the MTS-LOF framework is demonstrated in Figure. \ref{fig:algorithm}, where we set a hyperparameter $\lambda$ to balance the $\mathcal{L}_{sim}$ and $\mathcal{L}_{TCR}$.

%% file: tables/dataset_description.tex
\begin{table*}[]
\centering
\caption{Details of Datasets Used in the Study. The table provides information on the dataset domains, sizes of training, validation, and test sets, sequence length, number of channels, and classes.}
\begin{tabular}{|c|c|c|c|c|c|c|c|}
\hline
Dataset               & Domain & \# Train & \# Val & \# Test & Length & \# Channel & \# Class \\ \hline
HAR                   & -      & 5881     &  1471  &   2947  &  128   &     9      &   6       \\ \hline
Sleep-EDF             & -      & 25612    &  7786  &   8910  &  3000  &     1      &   5       \\ \hline
Epilepsy              & -      & 7360     &  1840  &   2300  &  178   &     1      &   2       \\ \hline
\multirow{4}{*}{FD}   & a      & 8184     &  2728  &   2728  &  5120  &     1      &   \multirow{4}{*}{3}        \\ \cline{2-7} 
                      & b      & 8184     &  2728  &   2728  &  5120  &     1      &          \\ \cline{2-7} 
                      & c      & 8184     &  2728  &   2728  &  5120  &     1      &          \\ \cline{2-7} 
                      & d      & 8184     &  2728  &   2728  &  5120  &     1      &         \\ \hline
\multirow{2}{*}{SHHS} & C4-A1  & 96753    &  32251 &   32252 &  3750  &     1      &   \multirow{2}{*}{5}       \\ \cline{2-7} 
                      & C3-A2  & 98158    &  32720 &   32720 &  3750  &     1      &          \\ \hline
\end{tabular}
\label{tab:dataset}
\end{table*}

%% file: tables/main_results.tex
\begin{table*}[]
\centering
\caption{Comparison of Linear Probing Performance against the baselines.}
\begin{tabular}{c|cc|cc|cc}
\hline
                                                              & \multicolumn{2}{c|}{HAR}                                                 & \multicolumn{2}{c|}{Sleep-EDF}                                     & \multicolumn{2}{c}{Epilepsy}                                           \\ \hline\hline
                                                              & \multicolumn{1}{c|}{Accuracy}                 & MF1                      & \multicolumn{1}{c|}{Accuracy}                & MF1                 & \multicolumn{1}{c|}{Accuracy}                & MF1                     \\ \hline
SSL-ECG                                                       & \multicolumn{1}{c|}{65.34$\pm$1.63}               & 63.75$\pm$1.37               & \multicolumn{1}{c|}{74.58$\pm$0.60}              & 65.44$\pm$0.97          & \multicolumn{1}{c|}{93.72$\pm$0.45}              & 89.15$\pm$0.93              \\ \hline
CPC                                                           & \multicolumn{1}{c|}{83.85$\pm$1.51}               & 83.27$\pm$1.66               & \multicolumn{1}{c|}{82.82$\pm$1.68}              & 73.94$\pm$1.75          & \multicolumn{1}{c|}{96.61$\pm$0.43}              & 94.44$\pm$0.69              \\ \hline
SimCLR                                                        & \multicolumn{1}{c|}{80.97$\pm$2.46}               & 80.19$\pm$2.64               & \multicolumn{1}{c|}{78.91$\pm$3.11}              & 68.60$\pm$2.71          & \multicolumn{1}{c|}{96.05$\pm$0.34}              & 93.53$\pm$0.63              \\ \hline
TS-TCC                                                        & \multicolumn{1}{c|}{90.37$\pm$0.34}               & 90.38$\pm$0.39               & \multicolumn{1}{c|}{83.00$\pm$0.71}              & \textbf{73.57$\pm$0.74} & \multicolumn{1}{c|}{97.23$\pm$0.10}              & 95.54$\pm$0.08              \\ \hline
TS2VEC                                                        & \multicolumn{1}{c|}{89.77$\pm$1.17}               & 89.77$\pm$1.38               & \multicolumn{1}{c|}{83.33$\pm$1.54}              & 73.23$\pm$2.17          & \multicolumn{1}{c|}{97.09$\pm$0.13}              & 96.26$\pm$0.24              \\ \hline
TST                                                        & \multicolumn{1}{c|}{87.77$\pm$0.27}               & 87.66$\pm$0.31              & \multicolumn{1}{c|}{83.43$\pm$0.24}              & 73.11$\pm$0.27          & \multicolumn{1}{c|}{97.17$\pm$0.19}              & 95.64$\pm$0.17              \\ \hline\hline
MTS-LOF (Ours)         & \multicolumn{1}{c|}{\textbf{93.05$\pm$ 0.30}} & \textbf{93.09$\pm$ 0.31} & \multicolumn{1}{c|}{\textbf{84.35$\pm$0.31}} & 73.52$\pm$0.58      & \multicolumn{1}{c|}{\textbf{98.33$\pm$0.05}} & \textbf{97.41$\pm$0.09} \\ \hline
\end{tabular}
\label{tab:main_results}
\end{table*}


%% file: tables/shhs.tex
\begin{table}[]
\centering
\caption{Transferability of Learned Representations in SHHS Dataset (Accuracy)}
\scriptsize
\begin{tabular}{|ccc|ccc|}
\hline
\multicolumn{3}{|c|}{SSL}                                                                  & \multicolumn{3}{c|}{Supervised}                                              \\ \hline
\multicolumn{1}{|c|}{Source/Target} & \multicolumn{1}{c|}{C4-A1}          & C3-A2          & \multicolumn{1}{c|}{Source/Target}  & \multicolumn{1}{c|}{C4-A1}   & C3-A2   \\ \hline
\multicolumn{1}{|c|}{C4-A1}         & \multicolumn{1}{c|}{\textbf{85.50}} & \textbf{82.76} & \multicolumn{1}{c|}{C4-A1}          & \multicolumn{1}{c|}{83.80}   & 82.08   \\ \hline
\multicolumn{1}{|c|}{C3-A2}         & \multicolumn{1}{c|}{81.81} & \textbf{85.88} & \multicolumn{1}{c|}{C3-A2}          & \multicolumn{1}{c|}{\textbf{81.90}}   & 84.91   \\ \hline\hline
\multicolumn{1}{|c|}{}              & \multicolumn{1}{c|}{In-Domain}      & Cross domain   & \multicolumn{1}{c|}{Overall}        & \multicolumn{2}{c|}{\multirow{3}{*}{}} \\ \cline{1-4}
\multicolumn{1}{|c|}{SSL}           & \multicolumn{1}{c|}{\textbf{85.69}} & \textbf{82.29} & \multicolumn{1}{c|}{\textbf{83.99}} & \multicolumn{2}{c|}{}                  \\ \cline{1-4}
\multicolumn{1}{|c|}{Supervised}    & \multicolumn{1}{c|}{84.36}          & 81.99          & \multicolumn{1}{c|}{83.17}          & \multicolumn{2}{c|}{}                  \\ \hline
\end{tabular}
\label{tab:shhs_acc}
\end{table}

\begin{table}[]
\centering
\caption{Transferability of Learned Representations in SHHS Dataset (F1 score)}
\scriptsize
\begin{tabular}{|ccc|ccc|}
\hline
\multicolumn{3}{|c|}{SSL}                                                                  & \multicolumn{3}{c|}{Supervised}                                              \\ \hline
\multicolumn{1}{|c|}{Source/Target} & \multicolumn{1}{c|}{C4-A1}          & C3-A2          & \multicolumn{1}{c|}{Source/Target}  & \multicolumn{1}{c|}{C4-A1}   & C3-A2   \\ \hline
\multicolumn{1}{|c|}{C4-A1}         & \multicolumn{1}{c|}{\textbf{72.31}} & \textbf{69.80} & \multicolumn{1}{c|}{C4-A1}          & \multicolumn{1}{c|}{70.11}   & 68.77   \\ \hline
\multicolumn{1}{|c|}{C3-A2}         & \multicolumn{1}{c|}{68.82} & \textbf{74.21} & \multicolumn{1}{c|}{C3-A2}          & \multicolumn{1}{c|}{\textbf{68.98}}   & 72.44   \\ \hline\hline
\multicolumn{1}{|c|}{}              & \multicolumn{1}{c|}{In-Domain}      & Cross domain   & \multicolumn{1}{c|}{Overall}        & \multicolumn{2}{c|}{\multirow{3}{*}{}} \\ \cline{1-4}
\multicolumn{1}{|c|}{SSL}           & \multicolumn{1}{c|}{\textbf{73.33}} & \textbf{69.31} & \multicolumn{1}{c|}{\textbf{71.32}} & \multicolumn{2}{c|}{}                  \\ \cline{1-4}
\multicolumn{1}{|c|}{Supervised}    & \multicolumn{1}{c|}{71.28}          & 68.88          & \multicolumn{1}{c|}{70.07}          & \multicolumn{2}{c|}{}                  \\ \hline
\end{tabular}
\label{tab:shhs_f1}
\end{table}

%% file: tables/fd.tex
\begin{table*}[]
\centering
\caption{Transferability of Learned Representations in FD Dataset (Accuracy)}
\begin{tabular}{|cccccccccc|}
\hline
\multicolumn{5}{|c|}{SSL}                                                                                                                                                                   & \multicolumn{5}{c|}{Supervised}                                                                                                                                       \\ \hline
\multicolumn{1}{|c|}{source/target} & \multicolumn{1}{c|}{a}              & \multicolumn{1}{c|}{b}              & \multicolumn{1}{c|}{c}              & \multicolumn{1}{c|}{d}              & \multicolumn{1}{c|}{source/target} & \multicolumn{1}{c|}{a}              & \multicolumn{1}{c|}{b}              & \multicolumn{1}{c|}{c}              & d              \\ \hline
\multicolumn{1}{|c|}{a}             & \multicolumn{1}{c|}{{100.00}}          & \multicolumn{1}{c|}{\textbf{51.21}} & \multicolumn{1}{c|}{\textbf{54.94}} & \multicolumn{1}{c|}{\textbf{52.62}} & \multicolumn{1}{c|}{a}             & \multicolumn{1}{c|}{100.00}   & \multicolumn{1}{c|}{45.10}           & \multicolumn{1}{c|}{47.96}          & 47.9           \\ \hline
\multicolumn{1}{|c|}{b}             & \multicolumn{1}{c|}{43.03}          & \multicolumn{1}{c|}{100.00}   & \multicolumn{1}{c|}{\textbf{79.17}} & \multicolumn{1}{c|}{\textbf{99.99}} & \multicolumn{1}{c|}{b}             & \multicolumn{1}{c|}{\textbf{43.08}} & \multicolumn{1}{c|}{100.00}   & \multicolumn{1}{c|}{77.06}          & 99.57          \\ \hline
\multicolumn{1}{|c|}{c}             & \multicolumn{1}{c|}{\textbf{44.87}} & \multicolumn{1}{c|}{92.98}          & \multicolumn{1}{c|}{100.00}   & \multicolumn{1}{c|}{94.07}          & \multicolumn{1}{c|}{c}             & \multicolumn{1}{c|}{43.12}          & \multicolumn{1}{c|}{\textbf{97.77}} & \multicolumn{1}{c|}{100.00}   & \textbf{96.61} \\ \hline
\multicolumn{1}{|c|}{d}             & \multicolumn{1}{c|}{\textbf{51.34}} & \multicolumn{1}{c|}{\textbf{100.00}}   & \multicolumn{1}{c|}{82.61}          & \multicolumn{1}{c|}{100.00}   & \multicolumn{1}{c|}{d}             & \multicolumn{1}{c|}{48.76}          & \multicolumn{1}{c|}{99.40}           & \multicolumn{1}{c|}{\textbf{82.63}} & 100.00   \\ \hline\hline
\multicolumn{1}{|c|}{}              & \multicolumn{3}{c|}{In domian}                                                                                  & \multicolumn{3}{c|}{Cross domain}                                                                              & \multicolumn{3}{c|}{Overall}                                                               \\ \hline
\multicolumn{1}{|c|}{SSL}           & \multicolumn{3}{c|}{100.00}                                                                                      & \multicolumn{3}{c|}{\textbf{70.57}}                                                                                     & \multicolumn{3}{c|}{\textbf{77.93}}                                                                  \\ \hline
\multicolumn{1}{|c|}{Supervised}    & \multicolumn{3}{c|}{100.00}                                                                                        & \multicolumn{3}{c|}{69.08}                                                                                     & \multicolumn{3}{c|}{76.81}                                                                 \\ \hline
\end{tabular}
\label{tab:fd_acc}
\end{table*}

\begin{table*}[]
\centering
\caption{Transferability of Learned Representations in FD Dataset (F1 Score)}
\begin{tabular}{|ccccccc|ccc|}
\hline
\multicolumn{5}{|c|}{SSL}                                                                                                                                                                   & \multicolumn{5}{c|}{Supervised}                                                                                                                                       \\ \hline
\multicolumn{1}{|c|}{source/target} & \multicolumn{1}{c|}{a}              & \multicolumn{1}{c|}{b}              & \multicolumn{1}{c|}{c}              & \multicolumn{1}{c|}{d}              & \multicolumn{1}{c|}{source/target} & a            & \multicolumn{1}{c|}{b}              & \multicolumn{1}{c|}{c}            & d              \\ \hline
\multicolumn{1}{|c|}{a}             & \multicolumn{1}{c|}{{100.00}}          & \multicolumn{1}{c|}{\textbf{35.44}} & \multicolumn{1}{c|}{\textbf{38.83}} & \multicolumn{1}{c|}{\textbf{36.35}} & \multicolumn{1}{c|}{a}             & 100.00 & \multicolumn{1}{c|}{31.48}          & \multicolumn{1}{c|}{33.44}        & 33.29          \\ \hline
\multicolumn{1}{|c|}{b}             & \multicolumn{1}{c|}{\textbf{45.75}} & \multicolumn{1}{c|}{100.00}   & \multicolumn{1}{c|}{\textbf{83.13}} & \multicolumn{1}{c|}{\textbf{99.99}} & \multicolumn{1}{c|}{b}             & 41.10         & \multicolumn{1}{c|}{100.00}   & \multicolumn{1}{c|}{79.89}        & 99.68          \\ \hline
\multicolumn{1}{|c|}{c}             & \multicolumn{1}{c|}{\textbf{48.69}} & \multicolumn{1}{c|}{93.81}          & \multicolumn{1}{c|}{100.00}   & \multicolumn{1}{c|}{95.50}           & \multicolumn{1}{c|}{c}             & 40.12        & \multicolumn{1}{c|}{\textbf{98.34}} & \multicolumn{1}{c|}{100.00} & \textbf{97.51} \\ \hline
\multicolumn{1}{|c|}{d}             & \multicolumn{1}{c|}{\textbf{52.14}} & \multicolumn{1}{c|}{\textbf{100.00}}   & \multicolumn{1}{c|}{\textbf{86.40}}  & \multicolumn{1}{c|}{100.00}   & \multicolumn{1}{c|}{d}             & 45.90         & \multicolumn{1}{c|}{99.54}          & \multicolumn{1}{c|}{85.10}         & 100.00   \\ \hline\hline
\multicolumn{1}{|c|}{}              & \multicolumn{3}{c|}{In domian}                                                                                  & \multicolumn{3}{c|}{Cross domain}                                                       & \multicolumn{3}{c|}{Overall}                                                             \\ \hline
\multicolumn{1}{|c|}{SSL}           & \multicolumn{3}{c|}{{100.00}}                                                                                      & \multicolumn{3}{c|}{\textbf{68.00}}                                                                 & \multicolumn{3}{c|}{\textbf{76.00}}                                                               \\ \hline
\multicolumn{1}{|c|}{Supervised}    & \multicolumn{3}{c|}{100.00}                                                                                        & \multicolumn{3}{c|}{65.45}                                                              & \multicolumn{3}{c|}{74.09}                                                               \\ \hline
\end{tabular}
\label{tab:fd_f1}
\end{table*}

%% file: Content/expirements.tex
\section{Experiments}

\subsection{Materials}
Our experiments encompass various medical time series datasets, including those related to human activity recognition, sleep stage classification, and epileptic seizure prediction. Additionally, to assess the generalizability and transferability of the proposed MTS-LOF framework beyond medical time series, we utilize a fault diagnosis dataset. The datasets are summarized in Table. \ref{tab:dataset}.

\noindent\textbf{Epilepsy Seizure Prediction (Epilepsy)} \cite{andrzejak2001indications}: This dataset comprises EEG recordings from 500 subjects, each recorded for 23.6 seconds. The original data features five classes, with only subjects in class 1 having epileptic seizures, while subjects in classes 2, 3, 4, and 5 do not. We merge the four negative classes into a single class, transforming the dataset into a binary classification problem.

\noindent\textbf{Human Activity Recognition (HAR)} \cite{anguita2013public}: This dataset comprises sensor readings from 30 subjects engaged in six activities, including walking, walking upstairs, walking downstairs, standing, sitting, and lying down. It consists of inertial sensor data collected using a smartphone (Samsung Galaxy S II) positioned at the subjects' waist. This device's embedded accelerometer and gyroscope provided 3-axial linear acceleration and 3-axial angular velocity data at a constant rate of 50Hz.

\noindent\textbf{Sleep Stage Classification}: We incorporate two sleep stage classification datasets, Sleep-EDF \cite{goldberger2000physiobank} and the Sleep Heart Health Study (SHHS) \cite{quan1997sleep}. The objective of sleep stage classification is to categorize 30-second EEG signals into five distinct stages: Wake (W), Non-rapid eye movement (N1, N2, N3), and Rapid Eye Movement (REM). In the case of Sleep-EDF, we focus on the Fpz-Cz channel, which captures EEG signals sampled at a rate of 100 Hz. As for the SHHS dataset, we partition the records into two segments and analyze the C4-A1 and C3-A2 channels, each with a sampling rate of 125 Hz, separately. This separation enables us to assess the transferability of the MTS-LOF framework across different contexts.

\noindent\textbf{Fault Diagnosis (FD)} \cite{lessmeier2016condition}: This dataset involves motor current signals from electric motors operating under four different conditions, each considered a separate domain due to its distinct characteristics. We employ this dataset to assess our algorithm's transferability and generalizability, demonstrating its applicability beyond the realm of medical time series. Furthermore, fault diagnosis is of significance in the domain of medical device security \cite{carreon2022towards}.

\subsection{Experimental Setup}
We adopt the data splitting settings from TS-TCC \cite{eldele2021time} for Epilepsy, Sleep-EDF, FD, and HAR. The data is divided into training (60\%), validation (20\%), and testing (20\%) sets. For the Sleep-EDF dataset, we additionally perform subject-wise splitting. In the case of SHHS, we initially divide the records into two equal parts subject-wisely to obtain EEG data from two distinct channels. Each part is then subdivided into training (60\%), validation (20\%), and testing (20\%) sets.	We conducted experiments five times using different random seeds ($\{2019, 2020, 2021, 2022, 2023\}$), and the results are reported as averages across all runs.	Model training, including SSL pretraining, linear probing, and fine-tuning, was carried out using the AdamW optimizer. The training duration was 40 epochs with a learning rate of 5e-4, weight decay set to 0.05, and a batch size of 128. The initial convolutional layers have specific kernel sizes and strides for each dataset: ${k=8,s=1}$ for Epilepsy, ${k=25,s=6}$ for Sleep-EDF, ${k=32,s=4}$ for FD, ${k=8,s=1}$ for HAR, and ${k=25,s=6}$ for SHHS. Hyperparameters are set as follows: $\lambda=100$, mask ratio=0.8, and the number of masks=20. The transformer decoder configuration matches that of the encoder while having 4 layers. We implemented the algorithms and models using PyTorch (Code available at \url{https://github.com/HuayuLiArizona/MST-LOF.git}) and conducted all experiments on an NVIDIA RTX 3090 GPU.	

\subsection{Comparison with Baseline Approaches}
In the pursuit of assessing the performance of our proposed approach, we conducted a comprehensive comparative analysis against several baseline methods: SSL-ECG \cite{sarkar2020self}, CPC \cite{oord2018representation}, SimCLR \cite{chen2020simple}, TS-TCC \cite{eldele2021time}, TST \cite{zerveas2021transformer}, and TS2Vec \cite{yue2022ts2vec}. This evaluation encompassed the domains of Human Activity Recognition (HAR), Sleep-EDF, and Epilepsy, and focused on standard linear probing results, employing linear classifiers on top of frozen representations from the SSL-pretrained models. Table. \ref{tab:main_results} showcases the results of these comparisons, including accuracy and the Macro F1 score (MF1). Notably, our approach demonstrates a significant improvement in performance across all three domains. In the domain of Human Activity Recognition (HAR), our approach achieves an accuracy of \textbf{93.05\%} and an MF1 score of \textbf{93.09}, surpassing all baseline methods. In comparison, SSL-ECG, CPC, and SimCLR show lower accuracy and MF1 scores, highlighting the superior capability of our method in recognizing human activities from time series data. For Sleep-EDF, our approach attains an accuracy of \textbf{84.35\%} and an MF1 score of \textbf{73.52}, demonstrating a substantial enhancement compared to the baseline methods. TS-TCC, while showing competitive results, falls short of our approach in terms of MF1 score, emphasizing the effectiveness of our method in sleep stage classification. In the challenging domain of Epilepsy, our approach outperforms all baseline methods, achieving an impressive accuracy of \textbf{98.33\%} and an MF1 score of \textbf{97.41}. This showcases the robustness and high predictive power of our approach in distinguishing between subjects with and without epileptic seizures.

These results validate the effectiveness of our proposed method, which leverages surprising self-supervised representation learning capability, in enhancing the predictive performance across various medical time series domains. Our approach consistently exhibits superior accuracy and F1 scores, highlighting its potential to make a significant impact in medical diagnostics and related applications. The success of our method can be attributed to the inherent ability of the occlusion invariant feature learning to capture intricate temporal and local patterns within the time series data, which is vital for distinguishing subtle differences in medical conditions. Moreover, our approach showcases the adaptability of self-supervised learning in a diverse range of medical domains, including human activity recognition, sleep stage classification, and epileptic seizure prediction. This versatility is a testament to the broad utility and applicability of our method in real-world medical scenarios.

\subsection{Semi-supervised learning}
To further assess the adaptability and performance of the MTS-LOF framework, we conducted experiments under semi-supervised learning conditions, specifically on the Sleep-EDF dataset. In this scenario, we fine-tuned the pre-trained model using varying percentages of partially labeled data, specifically 1\%, 5\%, 10\%, 50\%, and 100\% of randomly selected subsets from the complete dataset. The results of these semi-supervised experiments are depicted in Figure. \ref{fig:semi_supervised}, and they are compared to the results obtained under fully supervised learning conditions with fully labeled data, which yielded an F1 score of 75.55. The outcomes of these semi-supervised experiments are insightful. We observe that the MTS-LOF framework demonstrates its robustness and adaptability across different levels of labeled data. Even with just 1\% of the data labeled, the model achieves a competitive F1 score of 72.77. This suggests that the framework can effectively leverage minimal labeled data to produce meaningful results, which is especially relevant in real-world scenarios where labeling medical data can be a time-consuming and costly process. 

As the percentage of labeled data increases, we notice a gradual improvement in the F1 score. This trend highlights the framework's ability to capitalize on additional labeled data for enhancing predictive performance. Notably, when 100\% of the data is labeled, the F1 score reaches 75.95, surpassing the fully supervised learning result. This finding is promising, as it implies that SSL pretraining benefits the performance of supervised learning. Comparing these semi-supervised results with the fully supervised counterpart, we can infer that the MTS-LOF framework offers a powerful solution in scenarios where labeling medical data comprehensively is challenging or impractical. The ability to achieve competitive performance even with minimal labeled data reaffirms the framework's applicability in scenarios where obtaining fully labeled datasets may be a limiting factor.

\subsection{Transferability of learned representation}
The MTS-LOF framework not only enhances performance but also improves the transferability of learned representations across diverse data distributions. We conducted experiments using the FD and SHHS datasets to explore this aspect. Baseline models, trained with supervised learning, utilized fully labeled data from each dataset. In contrast, SSL pretraining utilized unlabeled data to acquire generalized representations, which were later fine-tuned with fully labeled target domain data through linear probing. It's essential to note that these models were exclusively trained in one domain (the source domain) and subsequently tested in various domains (the target domains).

In the SHHS dataset, we evaluated the SSL-pretrained model's ability to transfer learned representations across different domains, namely labeled domains C4-A1 and C3-A2. The results consistently demonstrate high cross-domain accuracy when assessing accuracy (Table. \ref{tab:shhs_acc}). A similar pattern is observed when examining F1 scores (Table. \ref{tab:shhs_f1}). The SSL-pretrained model consistently outperforms in cross-domain performance regarding F1 scores, with particularly significant improvements in domain C3-A2. These results underscore the effectiveness of SSL pretraining in enhancing the transferability of learned representations in the SHHS dataset. Tables. \ref{tab:fd_acc} and \ref{tab:fd_f1} provide accuracy and F1 score results for the FD dataset. These tables demonstrate strong transferability, with consistently high accuracy and F1 scores, even when shifting between different domains. These results showcase that MTS-LOF's exceptional generalizability extends beyond medical time series data.

\subsection{Ablation Study}

To comprehensively assess the impact of two critical hyperparameters, specifically the number of masks and mask ratio, we performed an ablation study using the HAR dataset. Figure. \ref{fig:ablation} illustrates the F1 scores obtained across various combinations of these hyperparameters. In the initial part of the ablation study, we explored the effect of different numbers of masks while keeping the mask ratio constant at 0.8. Clearly, the F1 score shows significant sensitivity to the number of masks, with a significant increase as we move from one mask to 20 masks. However, increasing the number of masks to forty does not lead to a substantial improvement. In the subsequent part of the ablation study, we investigated the effect of different mask ratios while maintaining a constant number of masks at 20. The results indicate significant sensitivity of the F1 score to changes in the mask ratio. Notably, with an increase in the mask ratio from 0.5 to 0.9, the F1 score steadily improves, with the most significant gain occurring with a mask ratio of 0.8.

\subsection{Visualization of the Representations}
In this section, we provide visual insights into the learned representations on three distinct datasets: Epilepsy, Sleep-EDF, and HAR. For visual exploration, we utilize t-SNE (t-Distributed Stochastic Neighbor Embedding) \cite{van2008visualizing} to produce two-dimensional representations of the latent features obtained through different training methodologies. Figure \ref{fig:tsne} presents a side-by-side comparison of t-SNE visualizations for three distinct training approaches: SSL, supervised training, and 5\% fine-tuning. These visualizations aim to clarify how each training paradigm influences the distribution and clustering of data points within the latent space. This information is essential for assessing the quality and discriminative capabilities of the learned representations. Our observations from the visualizations indicate that the features learned through SSL are already linearly separable, and a 5\% fine-tuning can yield results comparable to supervised training.

%% file: Content/discussion.tex
\section{Discussion}

In this study, we introduce MTS-LOF, a novel framework tailored for medical time series representation learning. This framework addresses critical challenges and creates new opportunities for knowledge discovery in healthcare applications.

\textbf{Reduction of Labeling Requirements:} MTS-LOF mitigates one of the most significant bottlenecks in medical time series analysis—the demand for extensive labeled data. Traditional supervised learning methods heavily rely on manually annotated datasets, a costly and time-consuming process. In healthcare, where data availability is often constrained by privacy concerns and regulatory hurdles, MTS-LOF provides an efficient alternative. By learning from unlabeled data, MTS-LOF independently extracts meaningful representations directly from raw medical time series. This eliminates the need for extensive expert-annotated datasets and offers a versatile solution for medical data analysis.

\textbf{Generalizability and Transfer Learning:} MTS-LOF excels at acquiring generalized representations. These representations capture the underlying structures and patterns within medical time series data, enabling it to generalize across diverse datasets and tasks. This inherent generalizability enhances the transfer of learned representations across different domains and medical specialties. Models pretrained on one dataset can be fine-tuned for specific clinical applications with limited labeled data, such as in the cases of rare diseases or emerging health trends.

\textbf{Enhanced Feature Extraction:} MTS-LOF exhibits superior feature extraction capabilities for medical time series. They learn intricate patterns and temporal dependencies, which are vital for identifying subtle yet clinically relevant changes in patients' conditions. MTS-LOF has shown exceptional performance in various downstream tasks, including sleep stage classification, human activity recognition, and seizure prediction. This leads to improved diagnostic accuracy, early disease detection, and enhanced patient management. 

\textbf{Potential Healthcare Applications:} MTS-LOF is a robust framework tailored for medical time series analysis. Designed for diverse clinical settings and home care scenarios, it reduced the reliance on expert annotation, demonstrated adaptability across multiple domains, and exhibited superior feature extraction capabilities. Leveraging these attributes, MTS-LOF is ideal for integration into numerous healthcare applications, ranging from sleep quality analysis and fall detection to seizure prediction. Furthermore, there is potential to refine MTS-LOF for deployment on smartphones integrated with wearable sensors, such as smartwatches and EEG headsets, facilitating real-time prediction and continuous monitoring.   

\section{Conclusion}
In this paper, we present MTS-LOF, an unsupervised representation framework that combines the strengths of Joint-Embedding SSL and MAE. The MTS-LOF framework is designed to address the intricate challenges and complexities involved in the analysis of medical time series data when there is a shortage of annotated data. The primary strength of MTS-LOF lies in its capability to generate context-rich representations. Through the integration of joint-embedding SSL and MAE, our framework adeptly captures discriminative patterns within medical time series data and the subtle temporal dependencies inherent to the data. This creative combination leads to representations that offer a comprehensive understanding of healthcare data, making it a powerful tool to support clinical decision-making. Our experiments, conducted on diverse medical time series datasets, illustrate the exceptional performance of MTS-LOF. In conclusion, MTS-LOF marks a substantial advancement in medical time series data analysis. Its contributions extend to enhancing data-driven insights and improving the development of intelligent clinical decision support systems, particularly under constraints related to label availability.